%% file: main.tex
\documentclass[10pt,twocolumn,letterpaper]{article}

\usepackage{iccv}
\usepackage{times}
\usepackage{epsfig}
\usepackage{graphicx}
\usepackage[dvipsnames]{xcolor}
\usepackage{microtype}
\usepackage{subfigure}
\usepackage{booktabs} 
\usepackage{multirow}
\usepackage[normalem]{ulem}
\useunder{\uline}{\ul}{}
\usepackage{caption}
\usepackage{pdflscape}
\usepackage{rotating}
\usepackage[accsupp]{axessibility}

\usepackage{amsmath}
\usepackage{amssymb}
\usepackage{mathtools}
\usepackage{amsthm}
\input{math_definition.tex}

\usepackage{xspace}
\usepackage{enumitem}

\theoremstyle{plain}

\theoremstyle{definition}

\theoremstyle{remark}

\usepackage[pagebackref=true,breaklinks=true,letterpaper=true,colorlinks,bookmarks=false]{hyperref}

\iccvfinalcopy 

\def\iccvPaperID{9767} 

\ificcvfinal\fi

\begin{document}

\title{Unified Out-Of-Distribution Detection: A Model-Specific Perspective}

\author{Reza Averly\\
The Ohio State University\\
{\tt\small averly.1@osu.edu}
\and
Wei-Lun Chao\\
The Ohio State University\\
{\tt\small chao.209@osu.edu}
}

\maketitle
\ificcvfinal\thispagestyle{empty}\fi

\input{abs}

\input{intro}

\input{related}
\input{unified}
\input{framework}
\input{exp}

\input{exp_results}
\input{disc}

{\small
\section*{Acknowledgments}
This research is supported in part by NSF (IIS-2107077, OAC-2118240, and OAC-2112606) and Cisco Research. We are thankful for the computational resources of the Ohio Supercomputer Center.
}

{\small
\bibliography{main.bib}
\bibliographystyle{ieee_fullname}
}


\iftrue
\appendix
\onecolumn
\input{appendix}

\fi


\end{document}

%% file: math_definition.tex
\usepackage{amssymb}
\usepackage{amsmath,amsfonts}
\usepackage{amsopn}
\usepackage{bm} 
\usepackage{multirow}
\usepackage{flushend}
\usepackage{tabularx}
\newcommand{\Ours}{\method{MS-OOD Detection}\xspace}

\newcommand{\OursLong}{{Model-Specific Out-of-Distribution (MS-OOD) Detection}\xspace}

\newcommand{\vct}[1]{\boldsymbol{#1}} 

\newcommand{\ProbOpr}[1]{\mathbb{#1}}

\newcommand{\expect}[2]{%
\ifthenelse{\equal{#2}{}}{\ProbOpr{E}_{#1}}
{\ifthenelse{\equal{#1}{}}{\ProbOpr{E}\left[#2\right]}{\ProbOpr{E}_{#1}\left[#2\right]}}} 

\newcommand{\vw}{\vct{w}}

\newcommand{\sS}{\mathcal{S}}

\newcommand{\sU}{\mathcal{U}}

\newcommand{\eat}[1]{}
\newcommand{\method}[1]{\textsc{#1}}

%% file: abs.tex
\begin{abstract}
Out-of-distribution (OOD) detection aims to identify test examples that do not belong to the training distribution and are thus unlikely to be predicted reliably. Despite a plethora of existing works, most of them focused only on the scenario where OOD examples come from semantic shift (\eg, unseen categories), ignoring other possible causes (\eg, covariate shift). In this paper, we present a novel, unifying framework to study OOD detection in a broader scope. Instead of detecting OOD examples from a particular cause, we propose to detect examples that a deployed machine learning model (\eg, an image classifier) is unable to predict correctly. That is, whether a test example should be detected and rejected or not is ``model-specific''. We show that this framework unifies the detection of OOD examples caused by semantic shift and covariate shift, and closely addresses the concern of applying a machine learning model to uncontrolled environments. We provide an extensive analysis that involves a variety of models (\eg, different architectures and training strategies), sources of OOD examples, and OOD detection approaches, and reveal several insights into improving and understanding OOD detection in uncontrolled environments.

\end{abstract}

%% file: intro.tex
\section{Introduction}
\label{s_intro}

Equipping a model with the capability to identify ``what it does not know'' is critical for reliable machine learning. Take image classification as an example. While state-of-the-art neural network models~\cite{he2016deep_resnet,dosovitskiy2020image,huang2017densely,liu2021swin} could perform fairly well on ``in-distribution (ID)'' data that belong to the training distribution, their accuracy often degrades drastically when facing data with covariate shift (\eg, different image domains or styles)~\cite{wilson2020survey} or semantic shift (\eg, novel categories)~\cite{liang2017enhancing_odin}. 
It is thus crucial to detect data on which the models cannot perform well and reject them from being classified.

Out-of-distribution (OOD) detection~\cite{hendrycks2016baseline_msp}, which aims to identify test examples drawn from a distribution different from the training distribution, is a promising paradigm toward such a goal. OOD detection has attracted significant attention lately, with a plethora of methods being developed \cite{ood_survey,salehi2021unified}. However, most of them focus solely on detecting examples with semantic shift, which arguably limits their applicability in uncontrolled environments where other kinds of OOD examples (\eg, with covariate shift) may appear.

\begin{figure}
\centering
\includegraphics[width=1\linewidth]{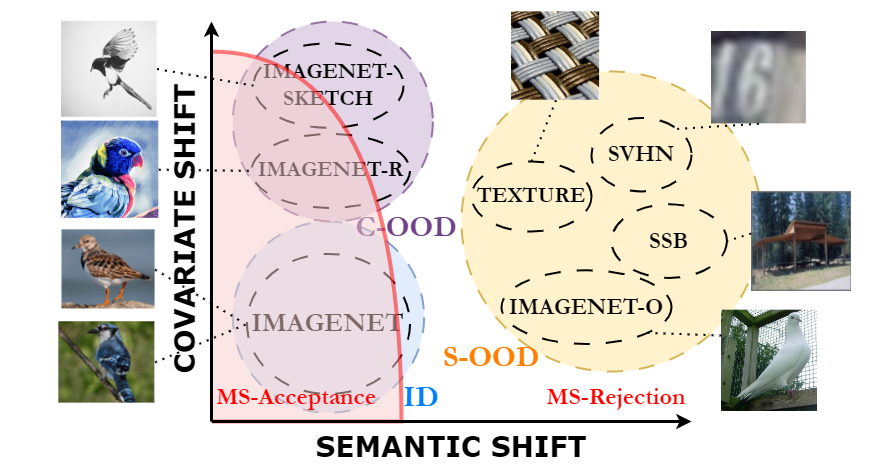}
\vskip -10pt
\caption{\small \textbf{\OursLong}, using ImageNet~\cite{russakovsky2015imagenet,imagenet_deng2009imagenet} as an example. The blue, purple, and yellow regions denote in-distribution (ID), covariate shift (C-OOD), and semantic shift (S-OOD) data, respectively, each with its datasets and representative images. Given an ImageNet classifier, the shaded red region denotes the \textit{correctly classified images} (called the \textit{acceptance region}). A robust classifier would have its acceptance region cover the ID and C-OOD data as much as possible. Using this framework, we can separate test data into \textbf{model-specific acceptance (MS-A)} and \textbf{rejection (MS-R)} cases, corresponding to the shaded 
 red region and its complement. 
The goal of \Ours is to detect the MS-R cases (\ie, examples misclassified by the classifier).}

\vskip -15pt
\label{figure_msood}
\end{figure}
 
In this paper, we thus attempt to expand the scope of OOD detection to further include covariate shift. At first glance, one may treat OOD examples with covariate shift (denoted as C-OOD) the same as OOD examples with semantic shift (denoted as S-OOD) --- \ie, to detect and reject as many of them as possible. However, unlike S-OOD examples, which the deployed classifier can never classify correctly, C-OOD examples have a chance to be correctly classified, as their label space is covered by the classifier. For instance, if the covariate shift is small (\eg, ImageNet~\cite{russakovsky2015imagenet} as ID; ImageNetV2 as C-OOD~\cite{imagenetv2_recht2019imagenet}), many of the C-OOD examples will likely be correctly classified. Even if the covariate shift is more pronounced (\eg, ImageNet-Sketch~\cite{imagenet_sketch_wang2019learning}), a robust classifier (\eg, with the CLIP-pre-trained backbone~\cite{radford2021learning}) is likely to still classify a decent amount of C-OOD examples correctly. Blindly rejecting these correctly classified C-OOD examples would adversely reduce the efficacy of the classifier.
 
Taking this insight into account, we propose a novel, unifying framework --- \textbf{\Ours} --- to study OOD detection from a ``\textbf{M}odel-\textbf{S}pecific'' perspective. In \Ours, \emph{whether an example should be detected and rejected from being classified (denoted by a ground-truth label $-1$) depends on whether the deployed classifier would misclassify it.} With this definition, every test example can be \emph{deterministically} assigned a ground-truth label based on the deployed classifier: $+1$ for correctly classified examples, which should not be rejected; $-1$ for misclassified examples, which should be rejected. This enables us to study different causes of OOD examples in a unifying way. It is worth noting that while C-OOD examples could be assigned different ground-truth labels, all the S-OOD examples are assigned ground-truth labels $-1$. In other words, similar to conventional OOD detection, all the S-OOD examples should be detected and rejected in \Ours. 

Nevertheless, unlike conventional OOD detection, which aims to accept all the ID examples drawn from the training distribution, \Ours, according to how it assigns ground-truth labels, aims to detect and reject misclassified ID examples as well. That is, what \Ours aims to accept are correctly classified ID and C-OOD examples; what \Ours aims to reject are misclassified ID and C-OOD examples, and all the S-OOD examples (which are always misclassified). Such a definition seamlessly unifies and generalizes the two related problems studied in the seminal work by Hendrycks and  Gimpel~\cite{hendrycks2016baseline_msp} --- detecting misclassified and OOD examples --- and could better reflect real-world application scenarios. For instance, for end-users who seek to reliably apply the machine learning model in uncontrolled environments, misclassification reveals the limitation of the model or the inherent difficulty of the examples (\eg, hard or ambiguous examples), implying the need for end-user intervention.

We conduct an extensive empirical study of \Ours. We consider three dimensions: \textbf{1) sources of OOD examples}, which include both semantic and covariate shift; \textbf{2) deployed classifiers}, which include different neural network architectures and training strategies; \textbf{3) OOD detection methods}, which include representative approaches such as Maximum Softmax Probabilities (MSP)~\cite{hendrycks2016baseline_msp}, Energy Score~\cite{liu2020energy}, Maximum Logit Score (MLS)~\cite{osr_goodclosedset}, Virtual-logit Matching (ViM)~\cite{wang2022vim}, and GradNorm~\cite{huang2021importance_gradnorm}. New datasets, classifiers, and OOD methods can easily be incorporated to broaden the scope. This experimental framework not only offers a platform to unify the community but also provides a ``manual'' to end-users for selecting the appropriate OOD methods in their respective use cases. 

Along with this study are 1) a list of novel insights into OOD detection and 2) a unifying re-validation of several existing but seemingly isolated insights found in different contexts. For instance, we find that the best detection methods for S-OOD, misclassified C-OOD, and misclassified ID data are not consistent; their effectiveness could be influenced by the paired model, hence ``model-specific''. 
Specifically for C-OOD examples, we find that the more robust the classifier is (\ie, having a higher accuracy in classifying C-OOD examples), the easier the misclassified C-OOD examples can be detected.
For S-OOD examples, while in general, we see the same trend as in~\cite{osr_goodclosedset} --- the stronger the classifier is (\ie, having a higher accuracy in classifying ID examples), the easier the S-OOD examples can be detected --- there are exceptions when we apply particular detection methods and classifiers. This suggests the need for a more thorough study. For misclassified ID examples, we find that they normally have lower scores (\eg, softmax probabilities) than correctly classified ID examples. In other words, one could set a higher threshold to reject more S-OOD examples without sacrificing the true positive rate of accepting ID examples that are correctly classified. Last but not least, we find that the baseline OOD detection method MSP~\cite{hendrycks2016baseline_msp} performs favorably in detecting misclassified ID and C-OOD examples, often outperforming other more advanced methods.

\noindent\textbf{Contributions.} Our contributions are two-folded:

\begin{itemize} [nosep,topsep=0pt,parsep=0pt,partopsep=0pt, leftmargin=*]
    \item We propose a novel framework, \Ours, which enables us to study different OOD examples (\eg, covariate shift and semantic shift) in a unifying way.
    \item We conduct an extensive study of \Ours, which reveals novel insights into OOD detection and unifies  existing insights gained from different contexts.
\end{itemize}

%% file: related.tex
\section{Related Works}
\label{s_related}

\begin{figure*}
\centering
\includegraphics[width=1\linewidth]{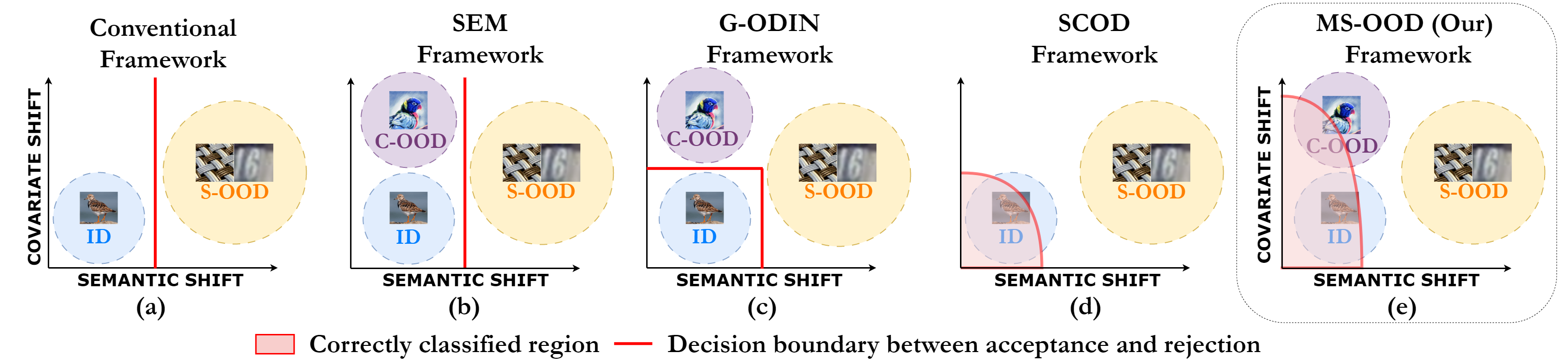}
\vskip -10pt
\caption{\small \textbf{Comparison of OOD detection frameworks}. The blue, purple, and yellow regions denote in-distribution (ID), covariate shift (C-OOD), and semantic shift (S-OOD) test data, respectively. The red boundary separates the test data into \textbf{acceptance (A)} and \textbf{rejection (R)} cases; the goal of OOD detection is to detect the R examples while accepting the A examples. Based on the test data involved and how they are separated, OOD detection frameworks can be categorized into (a) conventional OOD detection; (b) SEM~\cite{full_spectrum_ood_2022}; (c) G-ODIN~\cite{hsu2020generalized_odin}; (d) SCOD~\cite{augment_si_2022}; (e) our \Ours. It is worth noting that the red boundaries in (d) and (e) depend on the deployed multi-class classifiers trained for the ID classes --- the boundaries separate data that are correctly and wrongly classified by the classifier.}
\vskip -15pt
\label{figure_allframework}
\end{figure*}

\noindent\textbf{Out-of-distribution (OOD) detection settings.} OOD detection is highly related to anomaly detection, novelty detection, open-set recognition, and outlier detection~\cite{ood_survey}. The main differences lie in 1) the scope of OOD examples; 2) whether one has to perform multi-class classification on ID examples. Specifically, novelty detection and open-set recognition mainly consider OOD examples with semantic shift.

In conventional OOD detection, the focus is on detecting OOD examples with semantic shift (\ie, S-OOD), assuming the absence of covariate shift (\ie, C-OOD)~\cite{ood_survey,salehi2021unified}. Please see \autoref{figure_allframework} (a) for an illustration. Very few works include examples with covariate shift in their studies; most of them \emph{treat these examples as ID}, aiming to classify them robustly instead of detecting them as OOD~\cite{semantic_coherent_2021, spurious_2021, full_spectrum_ood_2022} (see \autoref{figure_allframework} (b), the ``SEM framework''~\cite{full_spectrum_ood_2022}). Some exceptions are~\cite{hsu2020generalized_odin, tian2021exploring_geometric_odin, wang2023how_sagarvaze_cood}, which aim to detect all examples with covariate shift as OOD (see \autoref{figure_allframework} (c), ``G-ODIN framework''~\cite{hsu2020generalized_odin}).

In anomaly detection and outlier detection, the focus is on differentiating ID and OOD examples without the need to classify ID examples (\ie, they treat ID examples as a single class).
Several works also consider examples with covariate shift~\cite{ood_survey}. Similar to the G-ODIN framework, these works aim to detect all examples with covariate shift as OOD.
 
We argue that whether an example with covariate shift should be detected or not depends on whether the deployed classifier would misclassify it or not. By taking a model-specific perspective, our \Ours resolves the dilemma between \textit{OOD detection}~\cite{ood_survey} and \textit{OOD generalization}~\cite{shen2021towards}: a robust model should \textit{generalizes} to examples with covariate shift; a weak model should \textit{reject} them.

\noindent\textbf{OOD generalization.}
Covariate shift is commonly studied in model generalization and robustness \cite{distrib_shift,shen2021towards}. Instead of rejecting examples with covariate shift, the community aims to improve the robustness of a neural network model so that the model could classify them correctly. However, given the notable accuracy gap between classifying ID examples and covariate-shift examples~\cite{ludwig_measure_robust}, we argue that it is desirable to also consider covariate shift in OOD detection. We note that a recent work~\cite{muller2023finding} also proposes to detect and reject covariate-shift examples if the shifts (measured by incompetence scores) are above a threshold.

\noindent\textbf{Selective classification.} Equipping a model with the option to reject has also been studied in selective classification \cite{geifman2017selective}. Different from OOD detection, selective classification focuses on rejecting uncertain ``ID'' examples. Recently, \cite{augment_si_2022} proposed to integrate selective classification with OOD detection (SCOD), aiming to detect both misclassified ID and semantic shift data (see \autoref{figure_allframework} (d)). In this context, our work can be seen as a generalized version, further taking covariate shifts into account (see \autoref{figure_allframework} (e)). Compared to \cite{augment_si_2022}, we provide a more comprehensive study, further emphasizing the role of models in evaluation.

\noindent\textbf{OOD detection methods} have roughly two categories: post-hoc and training-based~\cite{salehi2021unified}. 
The difference is whether one is allowed to specifically train the model (\eg, the classifier for ID data) to detect OOD examples. While training-based methods like outlier exposure~\cite{hendrycks2018deep_oe} have shown a much higher detection rate, they need access to ID training data and OOD examples that one would encounter, making them prohibitive for end-users and less effective in uncontrolled settings. 
We therefore focus on post-hoc methods. 
The baseline is to use the softmax output as confidence~\cite{hendrycks2016baseline_msp}. Other approaches consider scaling the temperature and adding input perturbations~\cite{liang2017enhancing_odin}; using logits~\cite{osr_goodclosedset}, energy~\cite{liu2020energy}, or gradients~\cite{huang2021importance_gradnorm} as the score; combining intermediate features with logits \cite{wang2022vim}.  

%% file: unified.tex
\section{Background} \label{unified_view}

We study the problem of out-of-distribution (OOD) detection in the context of multi-class classification. Given a neural network classifier $f$ that is trained on data sampled from a training distribution $P(X, Y\in\sS)$, OOD detection aims to construct a scoring function $g(x, f)$ which gives in-distribution (ID) data a higher score; OOD data, a lower score. $\sS$ here denotes the label space of the training data. 

During test time, examples that produce $g(x, f) > \tau$ are \emph{accepted} and forwarded for classification, while the rest are either \emph{rejected} or redirected for further investigation. Ideally, one would like to reject a test example $x$ if $f(x)\neq y$, where $y$ is the ground-truth class label. 
 
In real-world applications, sources of OOD examples can roughly be categorized into two groups: covariate shift and semantic shift. Using $P(X, Y)$ as in-distribution, these shifts occur either solely on the marginal distribution $P(X)$, or on both the class distribution $P(Y)$ and $P(X)$, respectively. In conventional experimental setups, semantic shift data consist of novel-class examples that are outside the classifier's label space $\sS$; covariate shift data consist of examples from the classifier's label space $\sS$ but from different domains.

%% file: framework.tex
\section{Model-Specific Out-of-Distribution Detection}

As introduced in~\autoref{s_intro} and~\autoref{s_related}, the focal point of existing OOD detection has been on semantic shift alone, assuming the absence of covariate shift during the test time. Very few works consider data with covariate shift in their studies, and there is a dilemma if one should detect those examples as OOD~\cite{hsu2020generalized_odin, tian2021exploring_geometric_odin, wang2023how_sagarvaze_cood} or accept them as ID~\cite{semantic_coherent_2021, spurious_2021, full_spectrum_ood_2022}. \autoref{figure_allframework} (a-c) illustrate these three frameworks.

In this paper, we argue that such a dilemma is more fundamental than previously believed. In essence, one ultimate goal of OOD detection is to detect test examples that the deployed classifier is unable to classify correctly and preclude them from being classified. To achieve such a goal, one must detect as many semantic shift examples (denoted as S-OOD) as possible, as they belong to novel classes and cannot be correctly classified by the deployed classifier. Covariate shift examples (denoted as C-OOD), in contrast, belong to the seen classes on which the deployed classifier was trained. Namely, some of the C-OOD examples are likely to be correctly classified; the amount of them depends on the degree of covariate shift and the robustness of the classifier. Taking this fact into account, we argue that the current dilemma in how to handle C-OOD examples can actually be resolved smoothly --- \emph{detecting and rejecting the misclassified C-OOD examples while accepting the correctly classified ones.}

\subsection{Problem definition}
\label{ss_problem_d}
In this subsection, we introduce \Ours, a novel framework that realizes the aforementioned goal and unifies the detection of OOD examples from different causes. \autoref{figure_msood} gives an illustration.

We consider a test environment that may contain ID, S-OOD, or C-OOD examples. Given a deployed classifier that was trained to classify the seen classes $\sS$ in the ID data, we separate the test examples into two cases:
\begin{itemize} [nosep,topsep=0pt,parsep=0pt,partopsep=0pt, leftmargin=*]
    \item \textbf{Model-specific acceptance (MS-A)}, which contains the ID and C-OOD examples that are correctly classified by the classifier, denoted by ID$+$ and C-OOD$+$, respectively. The red region in \autoref{figure_msood} shows this case.
    \item \textbf{Model-specific rejection (MS-R)}, which contains the ID and C-OOD examples that are misclassified by the classifier as well as all the S-OOD examples, denoted by ID$-$, C-OOD$-$, and S-OOD, respectively. The complement of the red region in \autoref{figure_msood} shows this case.
\end{itemize}

The goal of \Ours is to differentiate MS-A examples from MS-R examples, accepting the former while rejecting the latter.

\noindent\textbf{Remark.} 
Building upon \Ours, we can view the G-ODIN framework~\cite{hsu2020generalized_odin} and the SEM framework \cite{full_spectrum_ood_2022} as special cases under two model-specific assumptions. (Please see \autoref{figure_allframework} (c) and (b), respectively, for their illustrations.) The G-ODIN framework aims to reject all the C-OOD examples while accepting all the ID examples, equivalent to assuming that the deployed classifier performs perfectly on the ID data but cannot classify any C-OOD data correctly. The SEM framework aims to accept all the ID and C-OOD examples, equivalent to assuming that the deployed classifier can classify all these examples correctly. In essence, constructing such a classifier is the goal of model generalization and robustness, which is to maximize the red region in \autoref{figure_msood} to cover all the ID and C-OOD examples.

\subsection{Ground-truth label and detection mechanism}
\label{ss_sect4_detect}
We provide a more formal definition using notations introduced in~\autoref{unified_view}. Let $f: \mathcal{X} \mapsto \mathcal{S}$ be the classifier, where $\mathcal{S}$ is the model's label space. Given a labeled test example $(x\in\mathcal{X}, y\in \sS \cup \sU)$, where $\sU$ denotes the novel-class space, we assigns it a ground-truth \Ours label $z$:
\begin{itemize} [nosep,topsep=0pt,parsep=0pt,partopsep=0pt, leftmargin=50pt]
    \item \textbf{$z = +1$ for MS-A}, if $f(x) = y$;
    \item \textbf{$z = -1$ for MS-R}, if $f(x) \neq y$.
\end{itemize}

\Ours then aims to construct a scoring function $g(x, f)$ to predict $z$:
\begin{itemize} [nosep,topsep=0pt,parsep=0pt,partopsep=0pt, leftmargin=50pt]
    \item \textbf{$\hat{z} = +1$}, if $g(x, f) > \tau$;
    \item \textbf{$\hat{z} = -1$}, if $g(x, f) \leq \tau$,
\end{itemize}
where $\tau$ is the threshold (hyper-parameter) to be selected.

In this paper, instead of developing new scoring functions, we conduct an extensive study using existing OOD detection methods. More details are provided in~\autoref{s_exp}.

\subsection{Evaluation metric}
\label{ss_sect4_metric}
\begin{figure}
\centering
\includegraphics[width=1\linewidth]{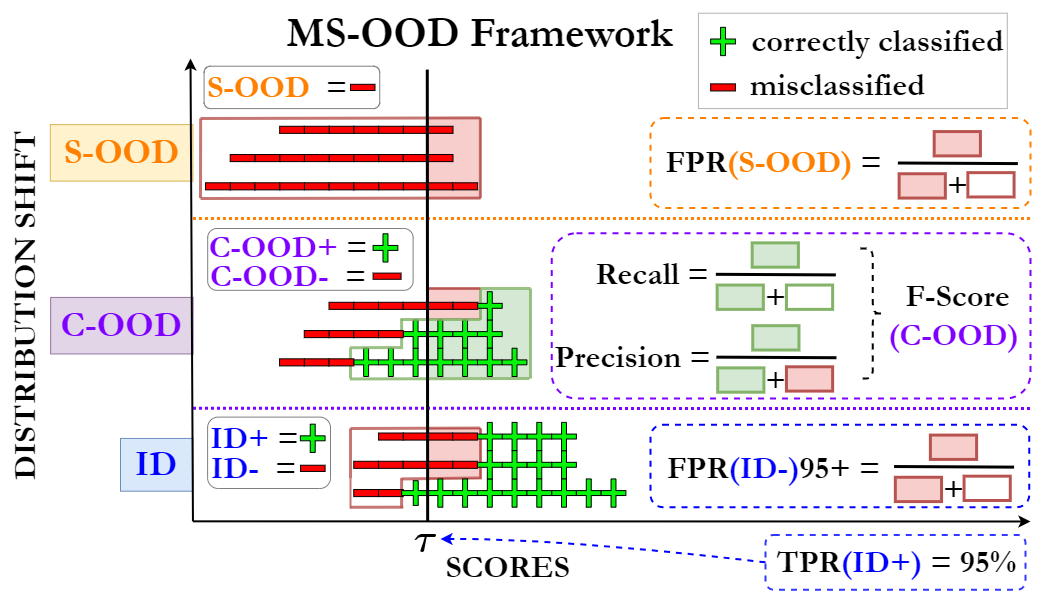}
\vskip -10pt
\caption{\small \textbf{Evaluation metrics for our \Ours.} {\color{green}$+$} and {\color{red}$-$}: correctly classified (MS-A) and misclassified (MS-R) examples by the deployed classifier $f$;  X-axis: the $g(x, f)$ score for \Ours; $\tau$: the threshold at TPR(ID+)$=95\%$. For MS-A examples ({\color{green}$+$}) that should be accepted, \ie, $g(x, f)>\tau$, the portion of them to the right of the threshold indicates the TPR. For MS-R examples ({\color{red}$-$}) that should be rejected, \ie, $g(x, f)\leq\tau$, such a portion indicates the FPR. For C-OOD examples, we further depict the Precision and Recall calculated at the threshold.}
\vskip -10pt
\label{figure_msood_metric}
\end{figure}

To evaluate \Ours in detail, we separately report the result on each ingredient of the MS-A (\ie, ID$+$, C-OOD$+$) and MS-R (\ie, ID$-$, C-OOD$-$, S-OOD) cases. Specifically, for MS-A examples that should be accepted by the scoring function (\ie, $g(x, f) > \tau$; $\hat{z}=+1$), we report the \textbf{True Positive Rate (TPR)}: the probability that examples with ground-truth labels $z=+1$ are predicted as $\hat{z}=+1$. For MS-R examples that should be rejected by the scoring function (\ie, $g(x, f) \leq \tau$; $\hat{z}=-1$), we report the \textbf{False Positive Rate (FPR)}: the probability that examples with ground-truth labels $z=-1$ are predicted as $\hat{z}=+1$. We use \textbf{$(\cdot)$} following these metrics to indicate the ingredient. For instance, \textbf{TPR(ID$+$)} is the TPR of ID$+$ examples; \textbf{FPR(S-OOD)} is the FPR of S-OOD examples.
 
As these metrics depend on the threshold $\tau$, we follow existing works to use one of them as the reference. Throughout the paper, we select $\tau$ such that it leads to \textbf{\color{Black}TPR(ID$+$)$=95\%$} unless stated otherwise. \textbf{{\color{Plum}FPR(S-OOD)}{\color{Black}@TPR(ID$+$)=95}} thus means the \textbf{{\color{Plum}FPR(S-OOD)}} at such a $\tau$ value.

We now list the metrics used in the main paper; \autoref{figure_msood_metric} gives an explanation. We include other metrics in the Suppl.

\begin{itemize} [nosep,topsep=0pt,parsep=0pt,partopsep=0pt, leftmargin=*]
    \item \textbf{{\color{Plum}FPR(S-OOD)}{\color{Black}@TPR(ID$+$)=95}}: FPR for accepting S-OOD data, at TPR$=95\%$ for accepting ID+ data. 
    \item \textbf{{\color{Plum}FPR(ID$-$)}{\color{Black}@TPR(ID$+$)=95}}: FPR for accepting ID$-$ data, at TPR$=95\%$ for accepting ID+ data. This is the same metric used in \cite{augment_si_2022} for selective classification.
    \item \textbf{{\color{Plum}F$_1$-Score(C-OOD)}{\color{Black}@TPR(ID$+$)=95}}: F$_1$-Score for identifying C-OOD$+$ data from C-OOD data, at TPR$=95\%$ for accepting ID+ data. 
\end{itemize}
We report F$_1$-Score as it simultaneously quantifies how well the C-OOD$+$ data are accepted and C-OOD$-$ data are rejected. We calculate Precision (P) and Recall (R) for identifying C-OOD$+$ data from C-OOD data, using $\tau$ that is selected to obtain TPR(ID$+$)$=95\%$. The F$_1$-Score is

\begin{align}
\frac{2\times P \times R} {P + R}.
\end{align}
\noindent\textbf{Remark.} \textbf{{\color{Plum}FPR(S-OOD)}{\color{Black}@TPR(ID$+$)=95}}, the metric we use to evaluate the detection of S-OOD examples, is very much the same as the one used in conventional OOD detection. The only difference is the threshold $\tau$: conventional OOD detection aims to accept all the ID examples and uses TPR(ID)=95 as the reference. 

\subsection{Comparison to SCOD~\cite{augment_si_2022}} Despite the superficial similarity to the SCOD framework \cite{augment_si_2022} shown in \autoref{figure_allframework} (d), our \Ours shown in \autoref{figure_allframework} (e) has several unique significance and contributions. First, as described in~\autoref{s_intro}, \Ours seeks a unifying framework to study both S-OOD and C-OOD examples, while SCOD combines selective classification with S-OOD detection. Second, we argue that the inclusion of C-OOD data is by no means trivial. As discussed in \cite{ood_survey}, existing OOD works do not agree on how to tackle C-OOD examples. Our ``model-specific'' perspective seamlessly glues the two extremes in the current dilemma (\ie, accepting or rejecting all C-OOD examples). Third, as depicted in \autoref{figure_allframework}, \Ours covers all the existing OOD detection frameworks. Some of them seem irrelevant or contrasting to each other at first glance, and \Ours unifies them: each of them can be viewed as a special case of \Ours under a certain assumption of the deployed model or the test environment (see~\autoref{ss_problem_d}). 
We consider this important as it provides a platform to connect different frameworks.

%% file: exp.tex
\section{Experimental Setup}
\label{s_exp}

Besides proposing \Ours, the other key contribution of the paper is a comprehensive empirical study and benchmarking. Specifically, we investigate three dimensions that could fundamentally influence the performance of \Ours: 1) sources of OOD examples, 2) deployed classifiers, and 3) OOD detection methods.

\subsection{Sources of ID/OOD examples}
\label{ss_dataset}
\noindent\textbf{In-distribution (ID).} We use the ImageNet-1K dataset~\cite{imagenet_deng2009imagenet} as the ID data. It is the standard benchmark for image classification and has been widely used as ID data in OOD detection. Many works in neural network architectures and training strategies also release their pre-trained classifiers.

\noindent\textbf{Semantic-shift (S-OOD).} We use the common benchmark datasets: SVHN \cite{netzer2011_svhn}, Texture (DTD) \cite{cimpoi14texture}, Places365 \cite{zhou2014places365}, iNaturalist \cite{van2018inaturalist} and SUN \cite{xiao2010sun}. 
We use the filtered versions proposed in \cite{huang2021mos} to ensure that classes in these datasets do not overlap with classes in ImageNet-1K.
{We also use the ImageNet-O dataset proposed in~\cite{imagenet_a_hendrycks2021nae}, which contains adversarially chosen classes to ImageNet-1K. Following~\cite{osr_goodclosedset}, we also consider the ``Easy'' and ``Hard'' subsets of ImageNet-21K~\cite{ridnik2021imagenet21k,imagenet_deng2009imagenet} proposed in the semantic-shift benchmark (SSB) for open-set recognition. These S-OOD data extracted from ImageNet-21K share similar styles with the ID data, more faithfully featuring {semantic} shift.}

\noindent\textbf{Covariate-shift (C-OOD)}. We consider four datasets with different degrees and types of covariate shift; their label spaces are covered by ImageNet-1K. ImageNetV2~\cite{imagenetv2_recht2019imagenet} is collected via a similar criterion as ImageNet-1K but after a decade, featuring a natural covariate shift by time.
For images with different styles and domains, we use ImageNet-R \cite{imagenet_r_hendrycks2021many} and ImageNet-S \cite{imagenet_sketch_wang2019learning} (``S'' stands for ``Sketch''). 
We also use
ImageNet-A \cite{imagenet_a_hendrycks2021nae}, a natural adversarial dataset curated by collecting wrongly predicted examples with high confidence. 
Since ImageNet-R and ImageNet-A contain only a subset of 200 classes from ImageNet-1K, when experimenting with them, we use the same subset of classes from ImageNet-1K as the ID data for a fair comparison. 

\subsection{Neural network models}
\label{ss_neuralnet}
Since \Ours is model-specific, we consider different neural network models for ImageNet-1K classification. 
We treat ResNet50~\cite{he2016deep_resnet} trained with the standard practice as the reference and consider factors that could influence 1) the classifier's strength in classifying the ID data and 2) its robustness in classifying the C-OOD data. Specifically, we consider \textbf{network depths} (\eg, ResNet18, ResNet152), \textbf{training strategies} (\eg, robust ResNet50~\cite{paszke2019pytorch} trained with the TorchVision new recipe\footnote{This involves data augmentation, label smoothing, longer training, etc. The resulting model has a higher ID and C-OOD classification accuracy.}), \textbf{pre-training} (\eg, CLIP-ResNet50~\cite{clip}), and \textbf{network architectures} (\eg, ViT-B-16~\cite{dosovitskiy2020image}). 
We obtain all these models from PyTorch \cite{paszke2019pytorch}, except for CLIP-ResNet50 which is released on its GitHub~\cite{clip}. 
We use the officially released ImageNet-1K version of CLIP, which leverages the zero-shot capability of CLIP without fine-tuning it on ImageNet-1K data: the final fully-connected layer is constructed by prompting with modified ImageNet class names. We note that despite employing a different training strategy, the robustness of CLIP largely comes from its diverse training data~\cite{fang2022data_determine_robust_clip}. Hence, we put the model under the umbrella of pre-training. 

\noindent\textbf{Evaluation metric.}
Besides the metrics introduced in~\autoref{ss_sect4_metric} for \Ours, we also report the multi-class classification accuracy of these models on the ID and C-OOD data to quantify their strength and robustness.

\subsection{Detection methods}
\label{ss_method}
We focus on post-hoc methods, assuming that the classifier is pre-trained and fixed. We argue that this better reflects the scenario of how end-users obtain and apply machine learning models.
We consider five representative methods categorized into output-based, feature-based, and hybrid. \emph{We use them to detect S-OOD, misclassified C-OOD, and misclassified ID examples.} Details are in the Suppl. 

\noindent\textbf{Output-based} methods. \textbf{Maximum Softmax Probabilities (MSP)} \cite{hendrycks2016baseline_msp} sets the scoring function $g(x, f)$ as the largest softmax probability outputted by the classifier $f$. MSP is treated as the baseline in most OOD detection literature. Despite its simplicity, we provide reasons why this algorithm is worth exploring in \Ours: \cite{osr_goodclosedset, limits_ood} showed the superior performance when MSP is paired with a strong classifier; \cite{augment_si_2022} showed the best performance compared to other state-of-the-art methods on {rejecting misclassified ID examples}; \cite{ming2022delving_mcm_clip} showed relatively good performance when MSP is paired with CLIP. Besides MSP, we also consider \textbf{Maximum Logit Score (MLS)}~\cite{osr_goodclosedset} and \textbf{Energy Score}~\cite{liu2020energy} as $g(x, f)$: MLS uses the logit value before softmax; Energy uses the denominator of the softmax calculation. 

\noindent\textbf{Feature-based} methods. We employ \textbf{GradNorm}~\cite{huang2021importance_gradnorm} which relies on gradients and the penultimate layer.

\noindent\textbf{Hybrid} methods. \textbf{Virtual-logit Matching (ViM)}~\cite{wang2022vim} uses residual features and logits to produce the $g(x, f)$ score. Its inner workings are quite similar to Mahalanobis~\cite{lee2018simple_mahalanobis} but ViM has a better performance. It is worth noting that ViM requires access to the training data to calculate the score.

\noindent\textbf{Remark.} We provide reasons why we exclude other post-hoc methods: ReAct~\cite{sun2021react} has an extra hyperparameter that can impact the accuracy of the classifier, changing the \textit{acceptance region} in \autoref{figure_msood} and leading to an unfair comparison. ODIN~\cite{liang2017enhancing_odin} needs hyperparameter tuning for input perturbations (hence the knowledge of OOD data). We include ODIN without tuning in Suppl.

We also provide reasons why we apply the scores developed for S-OOD detection to detect misclassified C-OOD and ID examples. First, C-OOD examples can essentially be seen as S-OOD examples if one treats ``style/domain'' + ``class'' as the semantic label. Second, scores like MSP are commonly used in domain adaptation~\cite{zou2018unsupervised,chen2021gradual,yang2021deep,kumar2020understanding} to detect unconfident predictions on out-of-domain data. Third, as studied in~\cite{hendrycks2016baseline_msp,augment_si_2022}, scores like MSP are quite useful for identifying misclassified ID data.

%% file: exp_results.tex
\section{Experimental Results}

We present the main results. \emph{For clarity, we mainly use figures (\eg, scatter plots) and leave detailed tables in Suppl.}

\input{table1_revamp}

\subsection{Model-agnostic C-OOD detection is ``ill-posed''}
\label{ss_sect6_illposed}
We first justify the necessity to introduce \Ours. We investigate the two extremes: accepting or rejecting all the C-OOD examples. 
We consider two classifiers: baseline ResNet50 and CLIP-ResNet50. The CLIP model has been shown surprisingly robust to C-OOD examples~\cite{fang2022data_determine_robust_clip,miller2021accuracy}. \autoref{new-Table-1} summarizes the results, in which we report the Accuracy (ACC) of classifying C-OOD examples, and the False Positive Rate (FPR) if we apply MSP to reject all C-OOD examples. For simplicity, we select $\tau$ at TPR(ID)=95 without separating ID data into ID$+$ and ID$-$.

On the three C-OOD datasets (ImageNet-R, ImageNet-S, and ImageNet-A) that have notable covariate shift from ImageNet-1K, CLIP-ResNet50 achieves higher ACC than ResNet50, demonstrating its robustness. With that being said, the ACC is still far from $100\%$, implying the risk to accept all C-OOD examples. 
At the other extreme, except for baseline ResNet50 on the challenging ImageNet-A dataset, we see non-zero ACC for the other combinations of models and datasets, implying a waste if we reject all C-OOD examples.
Perhaps more interestingly, for CLIP-ResNet50 which achieves higher ACC on C-OOD data, rejecting C-OOD data becomes harder, as evidenced by the much higher FPR(C-OOD) values. These results suggest the necessity to take a model-specific perspective in handling C-OOD data: accepting the correctly classified ones (\ie, C-OOD$+$) while rejecting the misclassified ones (\ie, C-OOD$-$).

\begin{figure}
\centering
\includegraphics[width=1\linewidth]{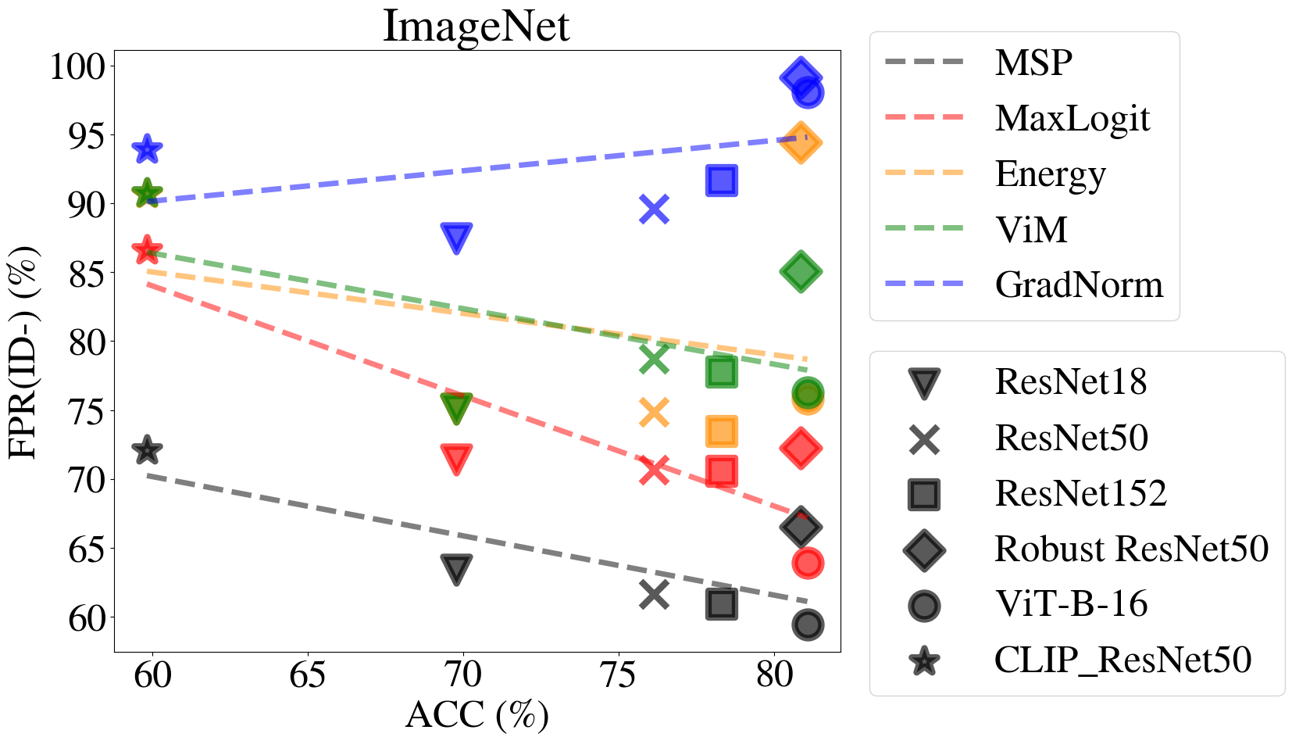}
\vskip -10pt
\caption{\small \textbf{ID in \Ours.} X-axis: ACC of classifying the ID examples; Y-axis: {\color{Plum}FPR(ID$-$)}{\color{Black}@TPR(ID$+$)=95} for wrongly accepting ID$-$ examples. We denote different neural network models by marker shapes; different detection methods by colors. The dashed line is the trend for each detection method.}
\label{fig_id_scatter}
\vskip -5pt
\end{figure}

\begin{figure}
\centering
\includegraphics[width=0.85\linewidth]{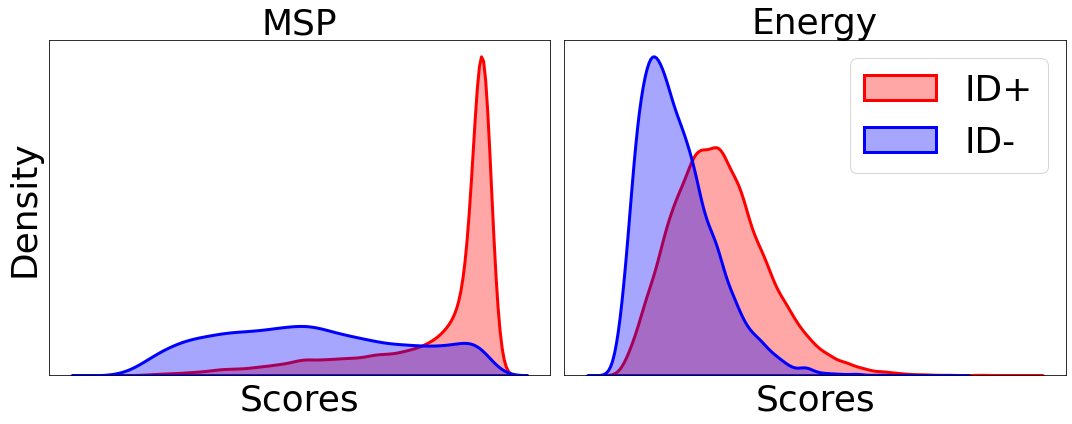}
\vskip -10pt
\caption{\small \textbf{Histogram of ID$+$ and ID$-$ data at different $g(x,f)$.} We use ResNet50 as the model $f$, and MSP and Energy for $g(x,f)$.}
\vskip -10pt
\label{fig_id_hist}
\end{figure}

\subsection{ID examples in \Ours}
\label{ss_exp_ID}
We now separately evaluate \Ours on different test data. We start with ID examples. We consider the four factors of models and the five detection methods introduced in~\autoref{s_exp}. The goal is to differentiate correctly classified and misclassified ID (\ie, ImageNet-1K validation) examples. We report \textbf{{\color{Plum}FPR(ID$-$)}{\color{Black}@TPR(ID$+$)=95}}.

\autoref{fig_id_scatter} summarizes the results, from which we have three key observations. First, generally speaking, the higher the ACC is in classifying ID data, the lower the FPR is in wrongly accepting ID$-$ examples. In other words, the stronger the classification model is, the easier it is to differentiate the correctly classified ID$+$ data from the remaining ID$-$ data.
Second, we find two exceptions. In particular, robust ResNet~\cite{paszke2019pytorch} ($\blacklozenge$) has a higher ACC than many other models but also a higher FPR (hence poor ID$-$ detection). Such an opposite trend also shows up when we apply GradNorm for ID$-$ detection: using a stronger model leads to a higher FPR. 
These exceptions suggest the need for a deeper look at 1) what model training strategies might hurt ID$-$ detection\footnote{\cite{osr_goodclosedset} is trained with data augmentation, label smoothing, etc.}, and 2) what underlying mechanisms of detection methods could not benefit from a stronger classifier. Specifically, we hypothesize that some of the training tricks may aggravate the overconfidence phenomenon of neural network predictions~\cite{guo2017calibration}.
Third, no matter which model is used, \textbf{MSP} outperforms other detection methods, achieving the lowest FPR. This aligns with what was reported in~\cite{augment_si_2022}. 

We further show the histogram of the ID$+$ and ID$-$ examples at different $g(x,f)$ values in~\autoref{fig_id_hist}. ID$+$ data usually have higher $g(x,f)$ than ID$-$ data. In other words, the $\tau$ value at TPR(ID$+$)=95 should be higher than the $\tau$ value at TPR(ID)=95. This implies that one could pick a higher threshold $\tau$ than what is normally set, to reject more S-OOD and C-OOD$-$ examples without sacrificing the TPR of accepting ID examples that are correctly classified.

\begin{figure}
\centering
\includegraphics[width=1\linewidth]{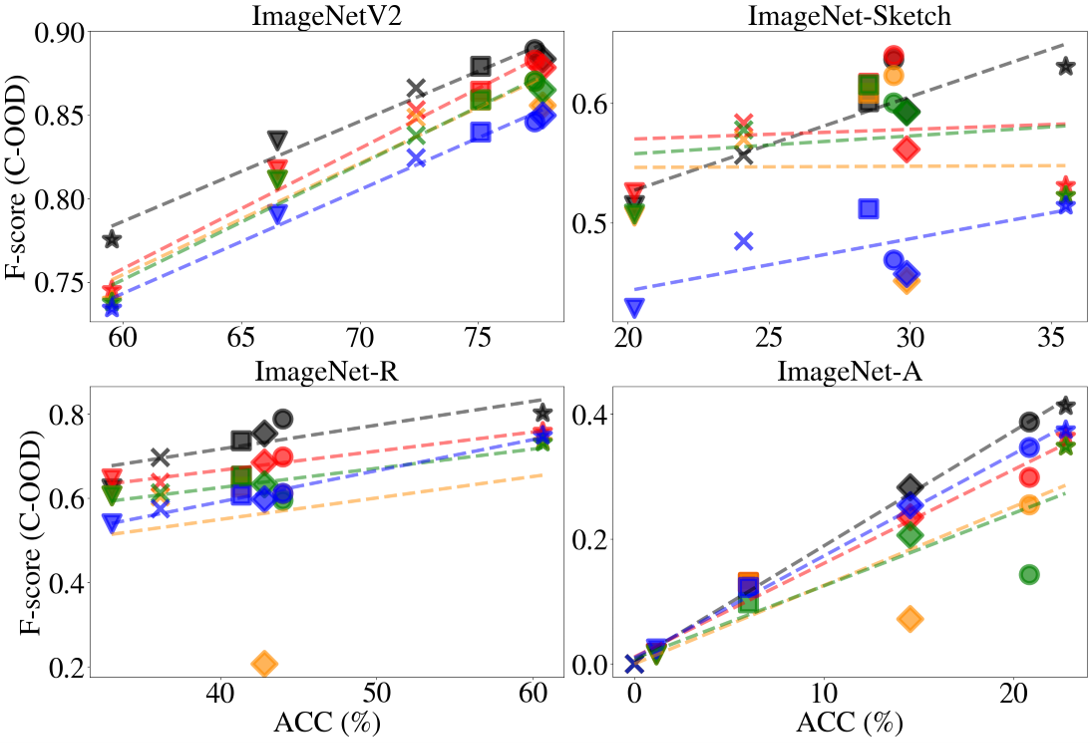}
\vskip -10pt
\caption{\small \textbf{C-OOD in \Ours.} Each sub-figure corresponds to each C-OOD dataset. X-axis: ACC of classifying the C-OOD examples; Y-axis: {\color{Plum}F$_1$-Score(C-OOD)}{\color{Black}@TPR(ID$+$)=95} for identifying C-OOD$+$ examples from C-OOD data. (The higher, the better.) Other denotations follow~\autoref{fig_id_scatter}.}
\label{fig_cood_scatter}
\vskip -5pt
\end{figure}

\begin{figure}
\centering
\includegraphics[width=0.8\linewidth]{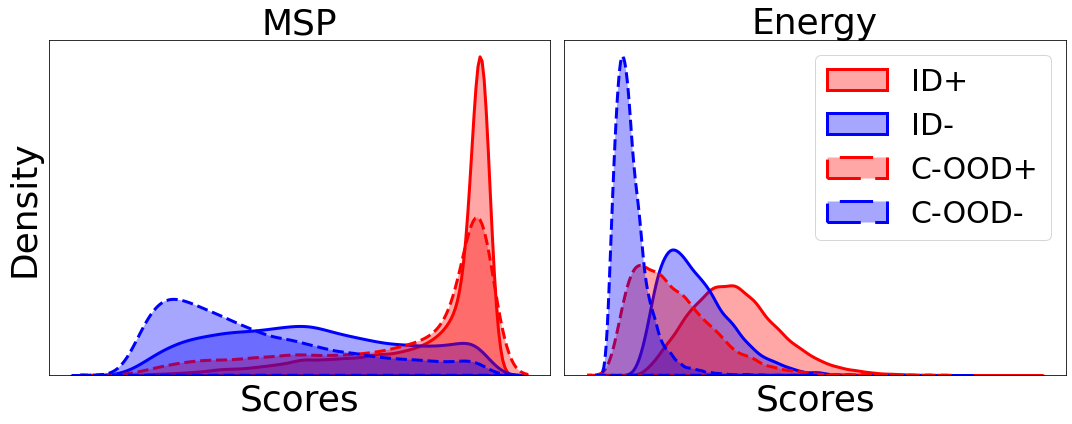}
\vskip -10pt
\caption{\small \textbf{Histogram of ID$+$, ID$-$, C-OOD$+$, and C-OOD$-$ data at different $g(x,f)$.} We use the ImageNet-R as C-OOD. We use ResNet50 as the model $f$, and MSP and Energy for $g(x,f)$.}
\vskip -10pt
\label{fig_cood_hist}
\end{figure}

\subsection{C-OOD examples in \Ours}
\label{ss_exp_COOD}
We now move on to C-OOD examples. We consider the four C-OOD datasets introduced in~\autoref{s_exp}. We report \textbf{{\color{Plum}F$_1$-Score(C-OOD)}{\color{Black}@TPR(ID$+$)=95}} defined in~\autoref{ss_sect4_metric}.

\autoref{fig_cood_scatter} summarizes the results, in which we draw a scatter plot for each C-OOD dataset. Different detection methods are marked by colors; different models are marked by shapes. The dashed line shows the trend for each detection method.

We have two key observations. First, in general, the higher the ACC is in classifying C-OOD data, the high the F$_1$-Score is in identifying C-OOD$+$ examples from C-OOD data. This can be seen across detection methods and across datasets. We view this a critical insight into bridging the dilemma of C-OOD detection and generalization --- \emph{improving the model's generalizability to C-OOD data would simultaneously benefit its ability to differentiate C-OOD$+$ data from C-OOD$-$ data.}
Second, like rejecting ID$-$ data, MSP generally performs the best (\ie, obtaining the highest F$_1$-Score) in differentiating C-OOD$+$ and C-OOD$-$ data.

We further show the histogram of the C-OOD$+$ and C-OOD$-$ examples, together with the ID$+$ and ID$-$ examples, at different $g(x,f)$ values in~\autoref{fig_cood_hist}. Overall, we find that it is typically simple to differentiate ID$+$ and C-OOD$-$ data. The challenge resides in how to differentiate C-OOD$+$ and ID$-$: accepting the former while rejecting the latter. We find that MSP could much better separate these two cases.

\begin{figure}
\centering
\includegraphics[width=1\linewidth]{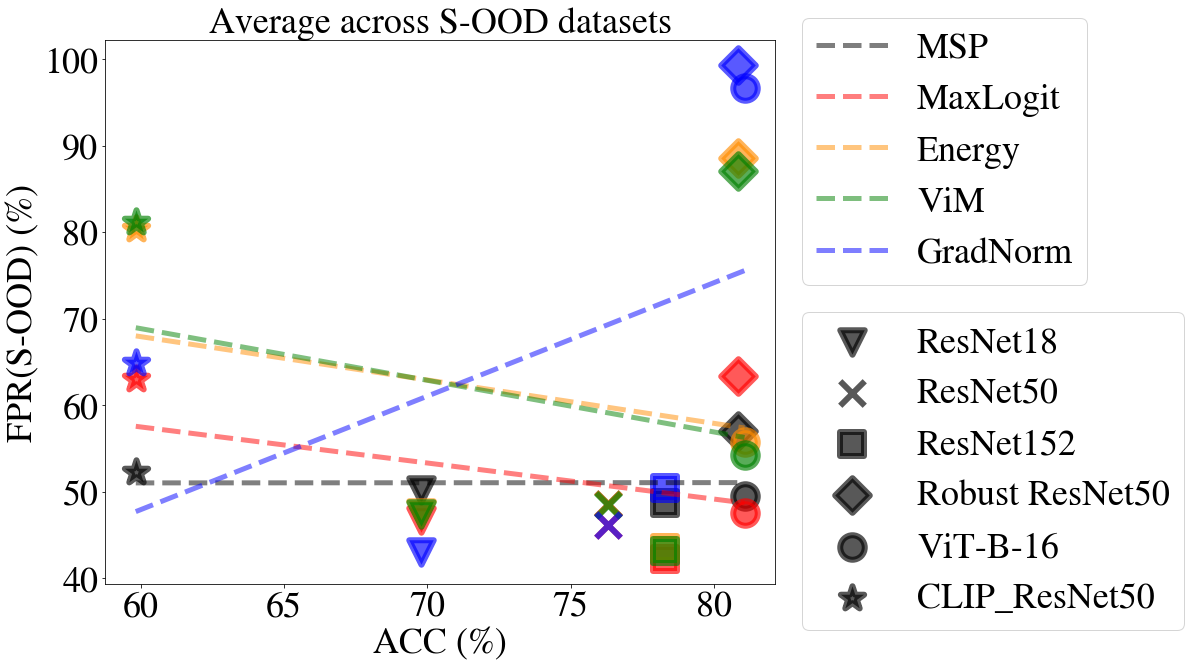}
\vskip -10pt
\caption{\small \textbf{S-OOD in \Ours.} The results are averaged over eight S-OOD datasets. X-axis: ACC of classifying the ID examples; Y-axis: {{\color{Plum}FPR(S-OOD)}{\color{Black}@TPR(ID$+$)=95}} for wrongly accepting S-OOD examples.} 
\label{fig_sood_scatter}
\vskip -5pt
\end{figure}

\begin{figure}
\centering
\includegraphics[width=1\linewidth]{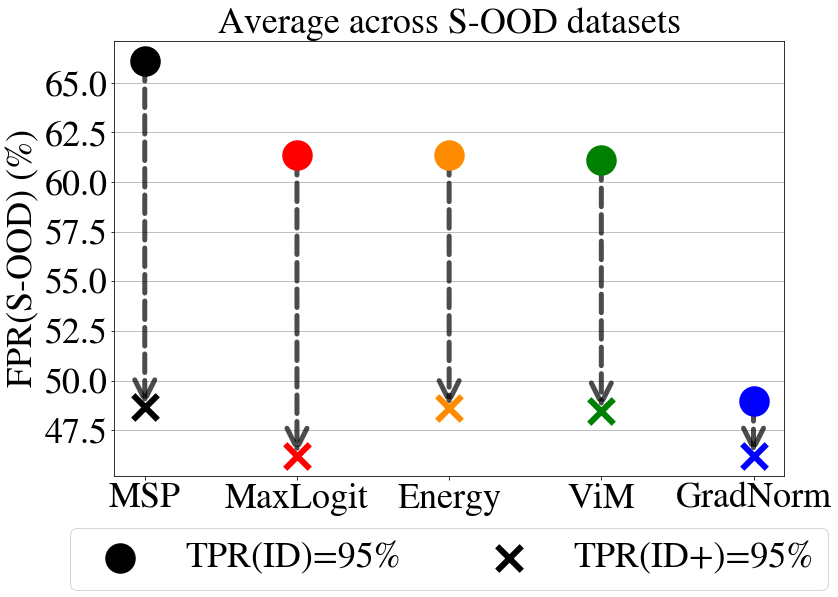}
\vskip -10pt
\caption{\small \textbf{Comparison of {\color{Plum}FPR(S-OOD)} at {\color{Black}TPR(ID$+$)=95}~($\times$) and {\color{Black}TPR(ID)=95}~($\circ$).} The main difference is the threshold $\tau$. We use the ResNet50 model; the results are averaged over eight S-OOD datasets. All detection methods improve at \color{Black}TPR(ID$+$)=95.}
\vskip -10pt
\label{fig_sood_comp}
\end{figure}

\subsection{S-OOD examples in \Ours}
\label{ss_exp_SOOD}
We now move on to S-OOD examples. We consider the eight S-OOD datasets introduced in~\autoref{s_exp}. We report \textbf{{\color{Plum}FPR(S-OOD)}{\color{Black}@TPR(ID$+$)=95}}.
\autoref{fig_sood_scatter} summarizes the results, in which we average the results over all datasets; the detailed results of individual datasets can be found in Suppl. Other setups and denotations exactly follow \autoref{fig_id_scatter}.

We have three observations. First, in general, the higher the ACC is in classifying ID data, the lower the FPR is in wrongly accepting S-OOD data. This is aligned with what was reported in~\cite{osr_goodclosedset}, suggesting that one could improve S-OOD detection by improving the deployed model. 
Second, we find similar exceptions as in~\autoref{ss_exp_ID}. When we apply GradNorm for S-OOD detection, using stronger models (hence higher ACC on ID data) leads to higher FPR on S-OOD detection; when we apply robust ResNet~\cite{paszke2019pytorch} ($\blacklozenge$) as the model, it has higher ACC on ID data but higher FPR on S-OOD detection. Similarly to~\autoref{ss_exp_ID}, we suggest a deeper look at these opposite trends. Finally, comparing different detection methods, we see no definite winner. For instance, the baseline MSP seems to fall behind in the high ACC regime, but it achieves the lowest FPR when paired with CLIP-ResNet50 (\textbf{$\star$}), as also pointed out in~\cite{ming2022delving_mcm_clip}. The multiple intersections among dashed lines further suggest the ``model-specific'' nature in picking the appropriate detection methods for S-OOD examples.

To draw a comparison to conventional S-OOD detection, we report its standard metric \textbf{{\color{Plum}FPR(S-OOD)}{\color{Black}@TPR(ID)=95}}. This metric features accepting all ID examples even if some are misclassified. Overall, we have quite similar findings to those mentioned above (please see Suppl.). Further comparing the two metrics in~\autoref{fig_sood_comp}, we see a consistent improvement in S-OOD detection using our new metric: FPR drops for all detection methods. 
In essence, since ID$+$ data normally have higher $g(x,f)$ scores than ID$-$ data (cf.~\autoref{fig_id_hist}), replacing {\color{Black}TPR(ID)=95} with {\color{Black}TPR(ID$+$)=95} allows us to use a larger $\tau$ to reject S-OOD data. While this comes with the cost of rejecting more ID data, most of them are indeed ID$-$ and do not hurt to be rejected. Perhaps more interesting, among all detection methods, MSP improves the most when setting the threshold by {\color{Black}TPR(ID$+$)=95}, arriving at the same level of FPR as other advanced methods. 

\begin{figure*}
\centering
\includegraphics[width=1\linewidth]{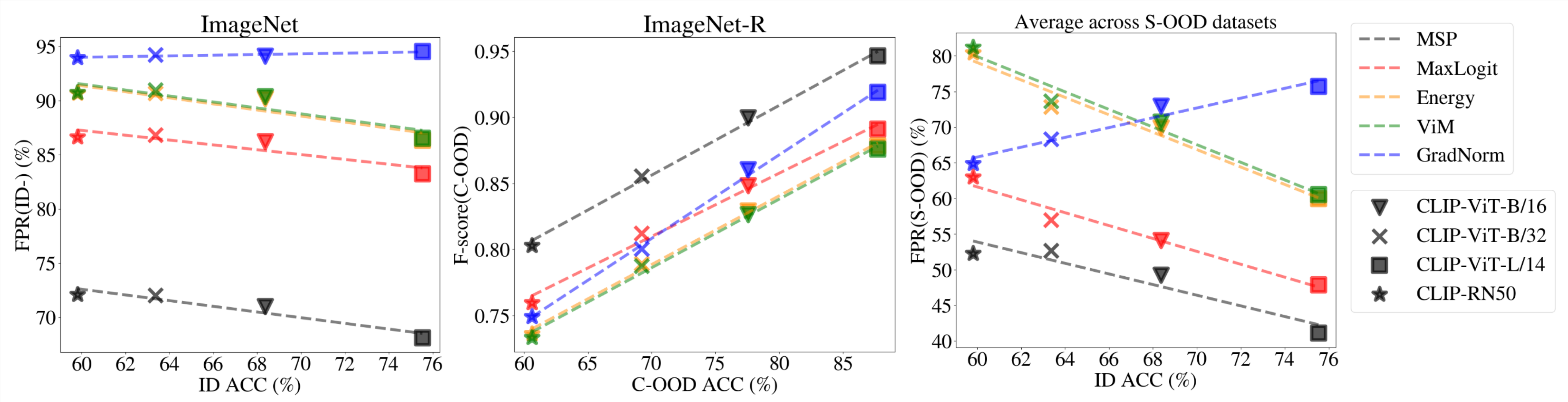}
\vskip -10pt
\caption{\small \textbf{Results using different CLIP backbones for ID$-$, C-OOD, and S-OOD in \Ours}. We use ImageNet-R for C-OOD and averaged over S-OOD datasets for S-OOD. Denotations follow \autoref{fig_id_scatter}, \autoref{fig_cood_scatter}, and \autoref{fig_sood_scatter}.}
\vskip -10pt
\label{fig_clip_exp}
\end{figure*}

\subsection{{Additional experiments} on CLIP~\cite{clip}}

\input{table_clip}

CLIP~\cite{clip} has shown superior robustness in many image classification tasks~\cite{fang2022data_determine_robust_clip}. We, therefore, further analyze whether our results generalize to CLIP with different backbone architectures. For this experiment, we include other pre-trained backbones such as ViT-B/16, ViT-B/32, and ViT-L/14~\cite{dosovitskiy2020image}. First, following \autoref{ss_sect6_illposed}, we investigate C-OOD detection in the extreme cases of accepting and rejecting all C-OOD examples.
We see a similar trend in~\autoref{tab:CLIP}: the higher the accuracy is in classifying C-OOD examples (\ie, the more robust the model is), the harder it is to reject these examples. These results suggest the necessity to take a model-specific perspective in handling C-OOD data.

Second, we investigate the performance in ID$-$, C-OOD (ImageNet-R), and S-OOD examples.
\autoref{fig_clip_exp} summarizes the results: we observe similar trends across all of them as in \autoref{ss_exp_ID}, \autoref{ss_exp_COOD}, and \autoref{ss_exp_SOOD}. 
These results suggest that our results apply to CLIP with different architectures. 

\subsection{More results in Suppl.}

Given the page limit, we leave additional results in Suppl., including the full tables that generate figures, results based on different metrics, and qualitative visualizations.

%% file: table1_revamp.tex
\begin{table}[]
\centering
\scriptsize
\caption{\small The accuracy (ACC) to classify C-OOD data and the false positive rate (FPR) to reject C-OOD data. For the former, we apply ResNet50 and CLIP-ResNet50. For the latter, MSP is used; $\tau$ is chosen for TPR(ID)=95. IN: ImageNet (see~\autoref{ss_dataset}).}
\tabcolsep 1pt
\vskip -10pt
\begin{tabular}{|l|cc|cc|cc|cc|cc}
\cline{1-9}
\multicolumn{1}{|c|}{\multirow{2}{*}{MODEL}} & \multicolumn{2}{c|}{IN-V2}                                                           & \multicolumn{2}{c|}{IN-S}                                                          & \multicolumn{2}{c|}{IN-R}                                                          & \multicolumn{2}{c|}{IN-A}                                                          & \multicolumn{2}{c}{} \\ \cline{2-9}
\multicolumn{1}{|c|}{}                       & \multicolumn{1}{c|}{ACC↑}  & \begin{tabular}[c]{@{}c@{}}FPR↓\\ (C-OOD)\end{tabular} & \multicolumn{1}{c|}{ACC↑} & \begin{tabular}[c]{@{}c@{}}FPR↓\\ (C-OOD)\end{tabular} & \multicolumn{1}{c|}{ACC↑} & \begin{tabular}[c]{@{}c@{}}FPR↓\\ (C-OOD)\end{tabular} & \multicolumn{1}{c|}{ACC↑} & \begin{tabular}[c]{@{}c@{}}FPR↓\\ (C-OOD)\end{tabular} &           &          \\ \cline{1-9}
RN50                                         & \multicolumn{1}{c|}{72.4} & 93.9                                                   & \multicolumn{1}{c|}{24.1} & 65.7                                                   & \multicolumn{1}{c|}{36.2} & 71.6                                                   & \multicolumn{1}{c|}{0.0}  & 81.5                                                   &           &          \\ \cline{1-9}
CLIP-RN50                                    & \multicolumn{1}{c|}{59.5} & 95.4                                                   & \multicolumn{1}{c|}{35.5} & 78.4                                                   & \multicolumn{1}{c|}{60.6} & 92.3                                                   & \multicolumn{1}{c|}{22.8} & 89.5                                                   &           &          \\ \cline{1-9}
\end{tabular}
\label{new-Table-1}
\vskip -10pt
\end{table}

%% file: table_clip.tex
\begin{table}[]
\centering
\scriptsize
\caption{\small The accuracy (ACC) to classify C-OOD data and the false positive rate (FPR) to reject C-OOD data. For the former, we apply CLIP-ResNet50, CLIP-ViT-B/16, CLIP-ViT-B/32, and CLIP-ViT-L/14. For the latter, MSP is used; $\tau$ is chosen for TPR(ID)=95. IN: ImageNet.}
\tabcolsep 1pt
\vskip -10pt
\begin{tabular}{|c|cc|cc|cc|cc|}
\hline
\multirow{2}{*}{MODEL} & \multicolumn{2}{c|}{IN-V2}                                                              & \multicolumn{2}{c|}{IN-S}                                                               & \multicolumn{2}{c|}{IN-R}                                                               & \multicolumn{2}{c|}{IN-A}                                                               \\ \cline{2-9} 
                       & \multicolumn{1}{c|}{ACC↑} & \begin{tabular}[c]{@{}c@{}}FPR↓\\      (C-OOD)\end{tabular} & \multicolumn{1}{c|}{ACC↑} & \begin{tabular}[c]{@{}c@{}}FPR↓\\      (C-OOD)\end{tabular} & \multicolumn{1}{c|}{ACC↑} & \begin{tabular}[c]{@{}c@{}}FPR↓\\      (C-OOD)\end{tabular} & \multicolumn{1}{c|}{ACC↑} & \begin{tabular}[c]{@{}c@{}}FPR↓\\      (C-OOD)\end{tabular} \\ \hline
CLIP-RN50              & \multicolumn{1}{c|}{59.5} & 95.4                                                        & \multicolumn{1}{c|}{35.5} & 78.4                                                        & \multicolumn{1}{c|}{60.6} & 92.3                                                        & \multicolumn{1}{c|}{22.8} & 89.5                                                        \\ \hline
CLIP-ViT-B/16          & \multicolumn{1}{c|}{68.8} & 94.8                                                        & \multicolumn{1}{c|}{48.2} & 83.7                                                        & \multicolumn{1}{c|}{77.6} & 94.1                                                        & \multicolumn{1}{c|}{50.1} & 89.7                                                        \\ \hline
CLIP-ViT-L/14          & \multicolumn{1}{c|}{75.8} & 94.6                                                        & \multicolumn{1}{c|}{59.6} & 87.4                                                        & \multicolumn{1}{c|}{87.7} & 96.1                                                        & \multicolumn{1}{c|}{70.7} & 91.2                                                        \\ \hline
\end{tabular}
\label{tab:CLIP}
\vskip-15pt
\end{table}

%% file: disc.tex
\section{{Discussion}}

\subsection{MSP performs well in ID$-$ \& C-OOD detection}
\label{ss_MSP_good}
Jointly comparing the results in \autoref{ss_exp_ID}, \autoref{ss_exp_COOD}, and \autoref{ss_exp_SOOD}, we find no winning detection method across all tasks. Nevertheless, the baseline MSP~\cite{hendrycks2016baseline_msp} performs quite well in detecting the ID$-$ examples and differentiating the C-OOD$+$ and C-OOD$-$ examples, outperforming other more advanced methods. High-levelly, it makes sense as MSP is optimized to avoid misclassification and has been widely used to detect unconfident data in domain adaptation. We provide more analyses in Suppl.

For example, \autoref{fig_id_cood_sood_hist_resnet50} and \autoref{fig_id_cood_sood_hist_clip_resnet50} show the score distributions: except for MSP, existing methods conflate misclassified data with correctly classified data on both ID and C-OOD.
Specifically on C-OOD, \autoref{table_cood_more} shows that MSP balances well between accepting correctly classified samples and rejecting misclassified examples.

\subsection{Improving performance in \Ours}
We discuss several directions.
First, building on the results in~\autoref{ss_exp_ID}, \autoref{ss_exp_COOD}, and \autoref{ss_exp_SOOD}, we suggest that one could improve the deployed classification model in terms of its accuracy and generalizability, as these mostly have positive correlations with the ID$-$, C-OOD, and S-OOD detection performance. 
Second, based on \autoref{ss_MSP_good}, we hypothesize that properly combining MSP and other OOD detection methods (\eg, through ensemble or cascade) or developing a new method that integrates the advantages of both would excel in \Ours. 

\section{Conclusion}

Out-of-distribution (OOD) detection has emerged as a promising topic for reliable machine learning. It has great potential to serve as a plug-and-play module to deployed machine learning models, helping them detect and reject potential misprediction. Nevertheless, looking at the literature, we found the topic seemingly overly fragmented into multiple scenarios with isolatedly developed algorithms~\cite{ood_survey,salehi2021unified}, despite that many of these algorithms share some common underlying principles (\eg, using logits or softmax probabilities). This could hinder the development of the topic and make it hard for end-users to apply the algorithms. In this paper, we make an attempt to unify the topic. We propose \Ours, a framework that aims to accept correctly predicted examples while rejecting wrongly predicted examples. We conduct an extensive study, considering in-distribution, semantic-shift, and covariate-shift test examples and different deployed models. We apply representative OOD detection methods and benchmark their performance in detail. Our study reveals new insights into OOD detection and unifies the existing ones. 

{Overall, our main contribution is a unifying framework to study OOD detection that involves semantic and covariate shifts. The novelty is in the framework, understanding, and insights. We deliberately do not propose a new OOD detection method but dedicate our paper to comprehensively evaluating existing methods, especially the post-hoc ones that are widely applicable}. We hope that our paper serves as a useful manual for OOD detection in practice.

%% file: appendix.tex
\section*{Supplementary Materials}
We provide additional information and results omitted in our main paper.
\begin{itemize}
    \item \textbf{Detailed setup (\autoref{supp_s_setup}):} including definitions of other metrics (cf.~\autoref{ss_sect4_metric}); descriptions of datasets (cf~\autoref{ss_dataset}) and OOD methods (cf~\autoref{ss_method}). 
    \item \textbf{Additional experimental results (\autoref{suppl_s_results}):} including detailed tables and figures for \autoref{ss_sect6_illposed}, \autoref{ss_exp_ID}, \autoref{ss_exp_COOD}, and \autoref{ss_exp_SOOD}.
\end{itemize}

\section{Detailed Setup}
\label{supp_s_setup}

\subsection{Metrics}
\label{suppl_s_metric}
\paragraph{Additional metrics.} In addition to the metrics defined in \autoref{ss_sect4_metric}, we report two other metrics for C-OOD examples:
\begin{itemize}
    \item \textbf{{\color{Plum}FPR(C-OOD$-$)}{\color{Black}@TPR(ID$+$)=95}}: FPR for accepting C-OOD$-$ data, at TPR$=95\%$ for accepting ID+ data. 
    \item \textbf{{\color{Plum}TPR(C-OOD$+$)}{\color{Black}@TPR(ID$+$)=95}}: TPR for accepting C-OOD$+$ data, at TPR$=95\%$ for accepting ID+ data.
\end{itemize}

\begin{figure}
\centering
\includegraphics[width=1\linewidth]{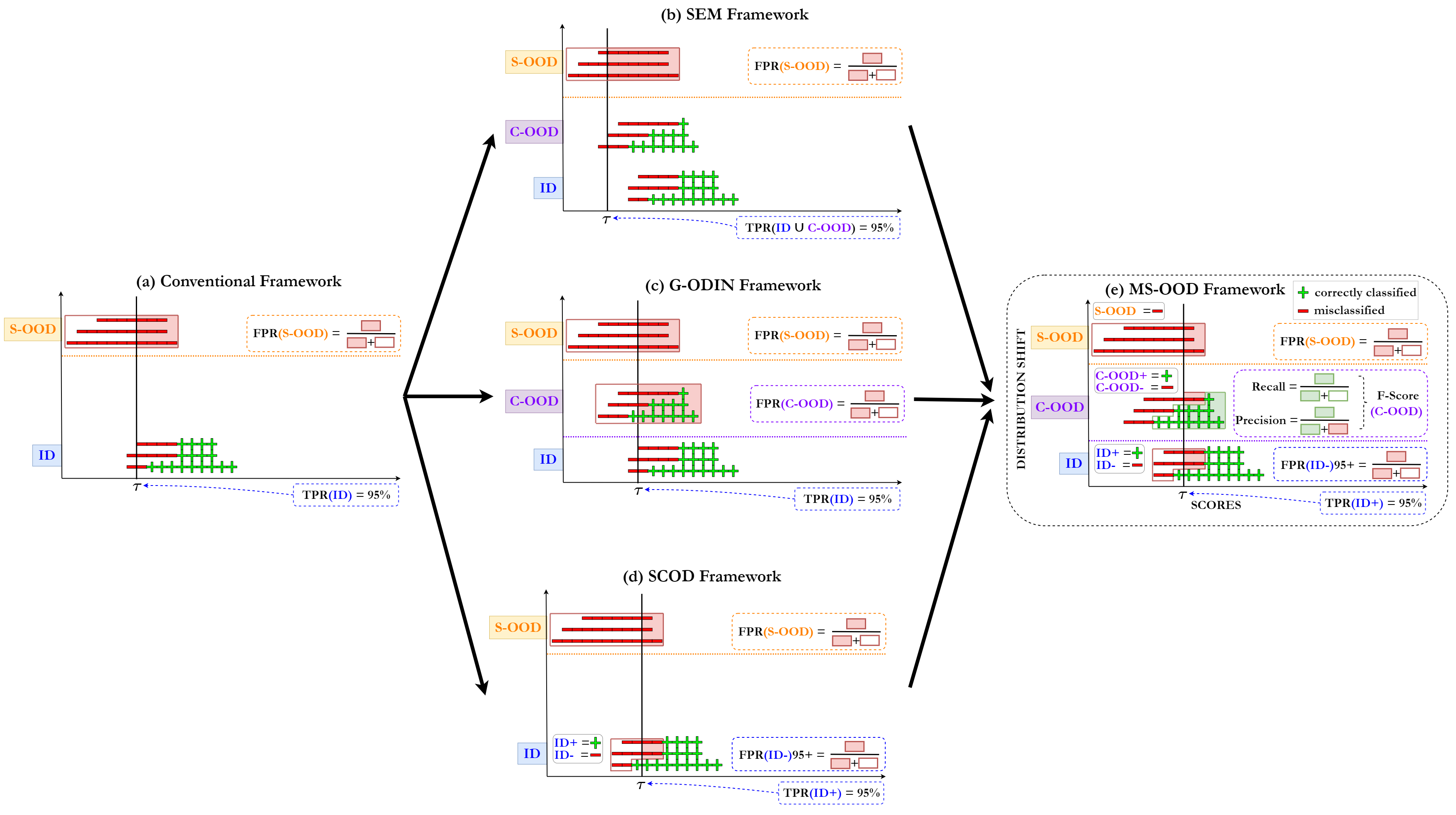}
\vskip -10pt
\caption{\small \textbf{Comparison of the evaluation metrics among (a) conventional, (b) SEM, (c) G-ODIN, (d) SCOD, and (e) \Ours frameworks}. Please be referred to \autoref{figure_msood_metric} for the denotations and \autoref{ss_sect4_metric} and \autoref{suppl_s_metric} for the definitions.}
\vskip -10pt
\label{figure_all_metrics}
\end{figure}

\paragraph{Metrics in other OOD detection frameworks.} To further distinguish between our~\Ours framework and the existing frameworks (cf.~\autoref{figure_allframework}), we describe their evaluation metrics in detail. We use the same notations described in \autoref{ss_sect4_metric}. \autoref{figure_all_metrics} contains a comprehensive visualization of the evaluation metrics of all frameworks. 

\begin{itemize}
    \item \textbf{Conventional framework (\autoref{figure_all_metrics} (a))} only considers the existence of ID and S-OOD data during testing. It evaluates the performance by FPR(S-OOD)@TPR(ID)=95: the False Positive Rate of wrongly accepting the S-OOD data when the True Positive Rate of the ID data is 95\%. Existing works in OOD detection often abbreviate it as FPR95.
    \item \textbf{SEM framework (\autoref{figure_all_metrics} (b))} extends the conventional framework by including C-OOD data. It evaluates the performance by the same metric; \ie, FPR(S-OOD) or FPR95. The difference lies in the threshold: SEM treats the C-OOD data as ID (since the model can potentially classify them correctly) and sets the threshold such that 95\% of the C-OOD and ID data are accepted. In this framework, if the scoring function is robust to covariate shift (\ie, the score distributions of the C-OOD and ID data are similar), the performance would be similar to the conventional framework. This is, however, not the case given that different models, covariate shifts, and detection methods can result in different score distributions between the ID and C-OOD data (see \autoref{fig_id_cood_sood_hist_resnet50} and \autoref{fig_id_cood_sood_hist_clip_resnet50}). 
    \item \textbf{G-ODIN framework (\autoref{figure_all_metrics} (c))} also includes C-OOD data during testing. However, instead of considering C-OOD data as ID, this framework aims to reject C-OOD data similarly to S-OOD data. It reports the FPR95 metric defined in the conventional framework separately for the S-OOD and C-OOD data, and we denote them by FPR(S-OOD) and FPR(C-OOD), respectively, where each is calculated when TPR for the ID data is 95\%. 
    \item \textbf{SCOD framework (\autoref{figure_all_metrics} (d))} incorporates selective classification into the conventional framework. That is, in addition to rejecting the S-OOD data, it further considers whether the model can classify the ID data correctly. The evaluation is based on FPR(S-OOD) and FPR(ID$-$) when TPR for the \textit{correctly classified} ID data (\ie, ID$+$) is 95\%. The differences to the conventional framework are two-fold: (1) rejection includes the misclassified ID data; (2) the threshold is based on the correctly classified ID data.
\end{itemize}

\paragraph{Metrics in open-set recognition.} Open-set recognition (OSR) tackles semantic shift detection and thus is highly related to conventional OOD detection. We remark that their metrics are relevant. In OOD detection, researchers often compare acceptance of ID data vs.~rejection of OOD data~\cite{ood_survey}. The open-set classification rate (OSCR) curve proposed in \cite{chen2021adversarial, dhamija2018reducing} shares similar concepts, comparing the classification accuracy of ID data vs. rejection of novel-class data. The key difference is that OOD detection implicitly assumes that all accepted ID data can be correctly classified, and our metric (\ie, TPR(ID$+$)) addresses it, making our metric more similar to the OSCR curve.

\subsection{Datasets}
We describe the datasets used in the main paper in detail.
\begin{itemize}
    \item \textbf{ImageNet-1k} (IN)~\cite{imagenet_deng2009imagenet} is a large-scale dataset for image recognition, containing 1,000 classes of real-world photos. It has
    1,281,167 training images, 50,000 validation images, and 100,000 test images. We use the validation set as the in-distribution (ID) test data.
    \item \textbf{ImageNet-V2} (IN-V2)~\cite{imagenetv2_recht2019imagenet} is collected similarly to ImageNet-1k with 3 test sets, each with 10,000 images. We use the test version that contains 10 images for each class with a selection frequency of at least 0.7.
    \item \textbf{ImageNet-R} (IN-R)~\cite{imagenet_r_hendrycks2021many} contains 30,000 images from 200 of the 1,000 ImageNet-1k classes. The images are 
    of different styles from the ImageNet-1k, including but not limited to art, cartoons, DeviantArt, graffiti, embroidery, graphics, origami, paintings, patterns, plastic objects, plush objects, sculptures, sketches, tattoos, toys, and video game renditions.
    \item \textbf{ImageNet-S} (IN-S)~\cite{imagenet_sketch_wang2019learning} consists of 50,000 sketch images; 50 images for each of the 1000 ImageNet-1k classes.
    \item \textbf{ImageNet-A} (IN-A)~\cite{imagenet_a_hendrycks2021nae} has 7,500 of natural adversarial images from 200 of the 1,000 ImageNet-1k classes. It is constructed based on the ResNet50 model's predictions. Wrongly classified images with high confidence are collected.
    \item \textbf{Street View House Numbers} (SVHN)~\cite{netzer2011_svhn} contains 73,257 training and 26,032 test images of digits 0 to 9. The images are 32 x 32 in resolution. We use only the test images for evaluation.
    \item \textbf{Describable Textures Dataset} (Texture)~\cite{cimpoi14texture} consists 5,640 textural images found in the wild with image resolution ranging from 300 x 300 to 640 x 640. It is split into training, validation and testing set with equal sizes. We use the training set for the evaluation. 
    \item \textbf{Places}~\cite{zhou2014places365} features more than 10 million images with more than 400 scene categories. We use the curated version from \cite{huang2021mos} with 50 classes outside ImageNet-1k classes and randomly sampled 10,000 images. 
    \item \textbf{iNaturalist} (iNat)~\cite{van2018inaturalist} consists of 859,000 images of more than 5,000 fine-grained species of animals and plants. We follow the setting from \cite{huang2021mos} with 110 classes not found in ImageNet-1k classes and randomly sample 10,000 images.
    \item \textbf{SUN}~\cite{xiao2010sun} has 130,159 images of 397 scenes with image resolution higher than 200 x 200. We also follow the same setting in \cite{huang2021mos} in this dataset, by using only 50 classes outside ImageNet-1k classes randomly sampling 10,000 images.
    \item \textbf{ImageNet-O} (IN-O)~\cite{imagenet_a_hendrycks2021nae} contains 7,500 natural adversarial images with classes outside a subset of 200 classes in ImageNet-1k. It is collected similarly to ImageNet-A.
    \item \textbf{Semantic Shift Benchmark} (SSB) \cite{osr_goodclosedset} is curated from ImageNet-21k~\cite{ridnik2021imagenet21k} by calculating the semantic distance using ImageNet-1k tree-like classes hierarchy. It is divided into 'Easy' (SSB-E) and 'Hard' (SSB-H) splits based on whether the classes are far way (\ie \textit{dog} with \textit{candle}) or close to ImageNet-1k classes (\ie \textit{dog} with \textit{wolf}). Each split has 1,000 classes and 50,000 images. 
\end{itemize}

\subsection{Detection methods}
Following the same notations introduced in \autoref{ss_sect4_detect}, we  briefly provide the mathematical definitions of the OOD detection algorithms used in this paper.

\paragraph{Maximum Softmax Probabilities (MSP)~\cite{hendrycks2016baseline_msp}} uses the softmax output of a classifier as the scoring function. Let us denote the model's label space by $\sS = \{1, 2, \cdots , C\}$, the model's output logit for class $c\in\sS$ by $f_c(x)$, and the training data by $D_\text{tr} = \{x_1, x_2, \cdots, x_N\}$. The scoring function for MSP is:
\begin{equation}\label{eq:msp}
g_{{\textstyle\mathstrut}\text{MSP}}(x, f) = \max_{c\in\sS} 
                \frac{e^{f_c(x)}}
                {\sum\limits^{C}_{c'=1}e^{f_{c'}(x)}}.
\end{equation}

\paragraph{Maximum Logit Score (MLS) \cite{osr_goodclosedset}} uses only the logits of the classifier:
\begin{equation}\label{eq:mls}
g_{{\textstyle\mathstrut}\text{MLS}}(x,f) = \max_{c\in\sS} f_c(x).
\end{equation}

\paragraph{Energy \cite{liu2020energy}} expresses the score by the denominator defined in \autoref{eq:msp}:
\begin{equation}\label{eq:energy}
g_{{\textstyle\mathstrut}\text{Energy}}(x, f) = T \cdot \log \sum_{c=1}^{C}e^{f_c(x)/T}, 
\end{equation}
where $T$ denotes the temperature. We follow \cite{liu2020energy} to set $T=1$ in the main paper.

\paragraph{Virtual-Logit Matching (ViM) \cite{wang2022vim}} introduces the Residual $\text{res}(x)$ to the energy-based method defined in \autoref{eq:energy}:
\begin{equation}\label{eq:vim}
g_{{\textstyle\mathstrut}\text{ViM}}(x,f) = \log \sum_{c=1}^{C}e^{f_c(x)} - \alpha\cdot\text{res}(x).
\end{equation}
The scaling parameter $\alpha$ can either be treated as a hyperparameter or computed using the formula:
\begin{equation}\label{eq:vim_alpha}
    \alpha =     \frac{\sum\limits_{i=1}^{N}\max\limits_{c\in\sS} f_c(x_i)}
        {\sum\limits_{i=1}^{N} \text{res}(x_i) }.
\end{equation}
We follow the same setting in \cite{wang2022vim}, using 200,000 uniformly sampled ImageNet training images to compute $\alpha$. Please be referred to~\cite{wang2022vim} for the detailed derivation of $\text{res}(x_i) $.

\paragraph{GradNorm \cite{huang2021importance_gradnorm}} relies on the gradients~w.r.t.~the last fully-connected layer of the classifier. 
Let us define the cross-entropy loss as:

\begin{equation}\label{eq:ce}
    \mathcal{L}_{\text{CE}} (f(x), {y}) 
        = - \log\frac{e^{f_{y}(x)}}{\sum\limits^{C}_{c=1}e^{f_c(x)}},
\end{equation}
where ${y}$ denotes the ground-truth label of $x$. The scoring function of GradNorm is defined as:
\begin{equation}\label{eq:gradnorm}
    g_{{\textstyle\mathstrut}\text{Grad}}(x, f) =\|\frac{1}{C} \sum_{c=1}^C\nabla_{\vw}\mathcal{L}_{\text{CE}} (f(x), c)\|_1,
\end{equation}

where $\vw$ denotes the weights (represented as a vector) of the last fully-connected layer. This score represents the average of the derivatives of cross entropy over all classes. 


\section{Additional Experimental Results}
\label{suppl_s_results}

\begin{figure}
\centering
\includegraphics[width=1\linewidth]{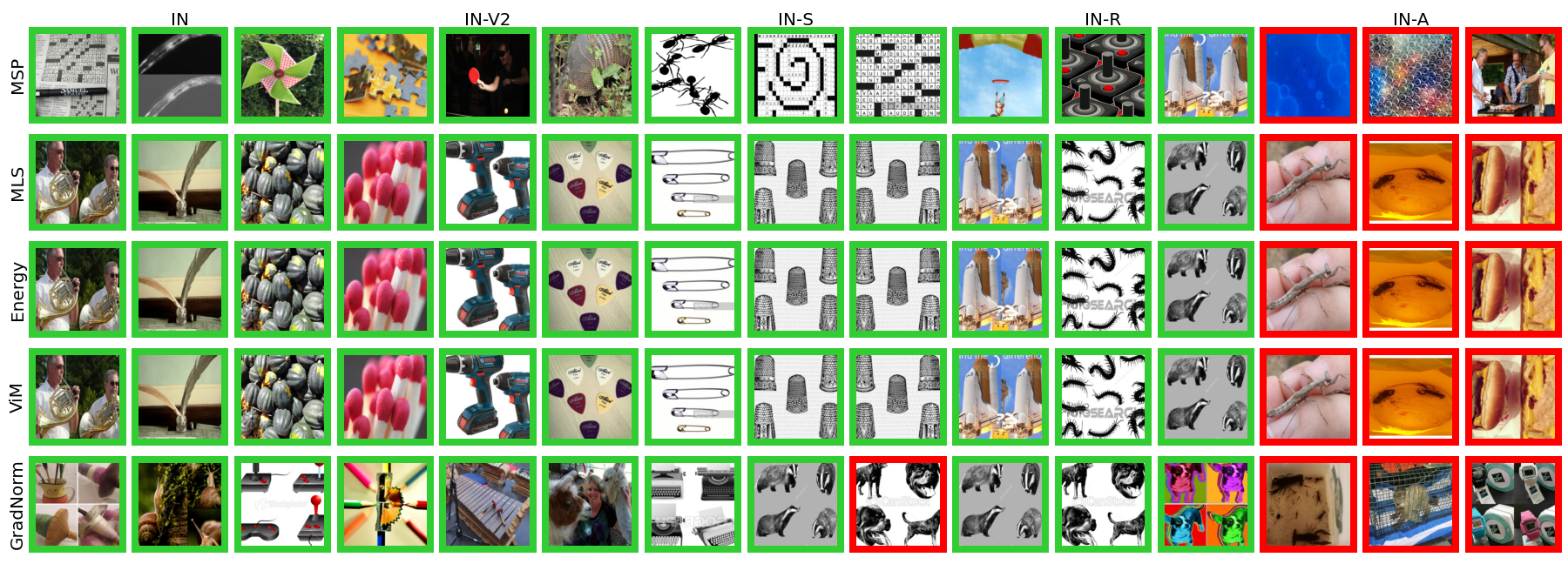}
\vskip -10pt
\caption{\small \textbf{Top 3 image examples for each ID and C-OOD datasets across different detection algorithms using ResNet50.} {\color{Red}Red} and {\color{green}green} border denote \textit{misclassified} and \textit{correctly classified} images consecutively. The row corresponds to detection algorithms (from top to bottom: MSP, MLS, Energy, ViM, GradNorm) and the column corresponds to datasets(from left to right for every three columns: ImageNet, ImageNet-V2, ImageNet-S, ImageNet-R, ImageNet-A).} 
\label{fig_resnet50_top}
\vskip -5pt
\end{figure}

\begin{figure}
\centering
\includegraphics[width=1\linewidth]{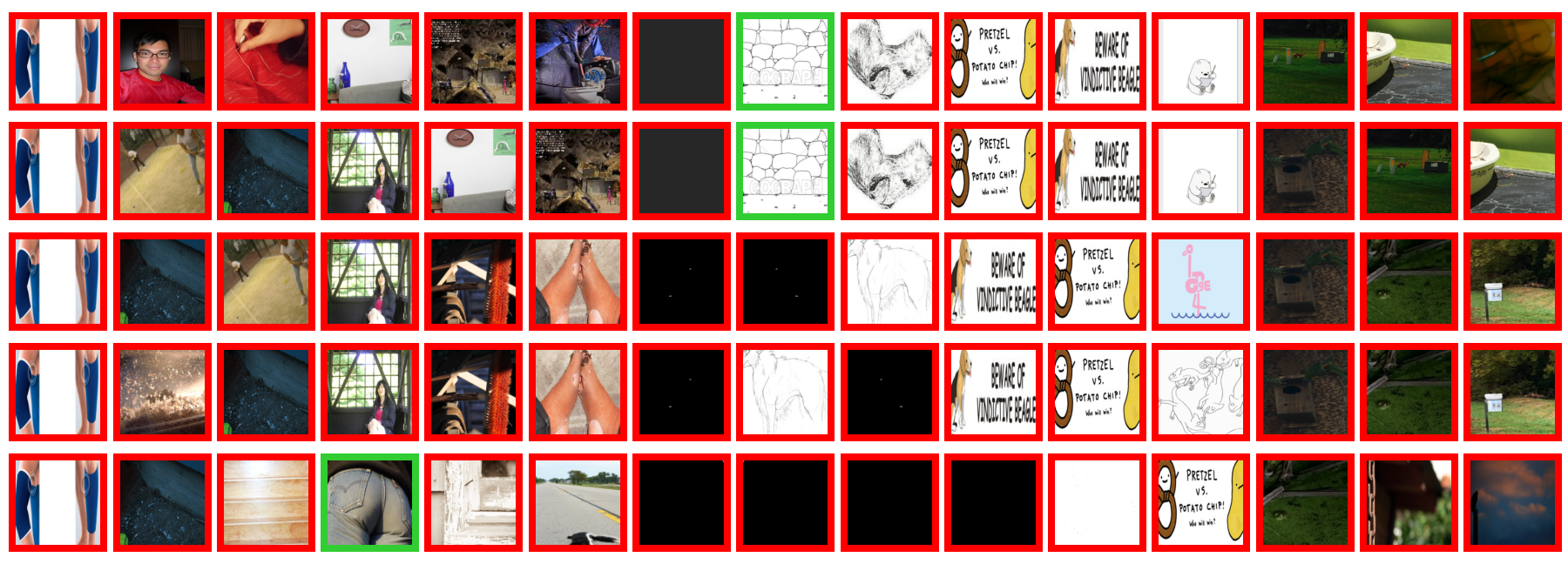}
\vskip -10pt
\caption{\small \textbf{Bottom 3 image examples for each ID and C-OOD datasets across different detection algorithms using ResNet50.} Denotations follow \autoref{fig_resnet50_top}.} 
\label{fig_resnet50_bottom}
\vskip -5pt
\end{figure}

\input{table_cood_ill}
\input{table_id_cood_sood}

\subsection{Qualitative visualization}
We show in~\autoref{fig_resnet50_bottom} and~\autoref{fig_resnet50_top} several examples of the C-OOD data and ID data. Specifically, we consider ResNet50 with different OOD detection algorithms and we sort the examples by the $g(x, f)$ scores: the higher the score is, the higher chance that it is to be accepted. We then group them into top 3 in \autoref{fig_resnet50_top} and bottom 3 \autoref{fig_resnet50_bottom} based on the scores and further indicate if they are correctly classified (\ie, ID$+$ or C-OOD$+$) or not (\ie, ID$-$ or C-OOD$-$). We find that across algorithms the most likely rejected and accepted images are consistent. Furthermore, examples with lower $g(x, f)$ scores~\autoref{fig_resnet50_bottom} are visually much more different (\ie mostly black or white images on ImageNet-S and ImageNet-R), suggesting why they are mostly classified incorrectly.

\subsection{Additional results for \autoref{ss_sect6_illposed}}

We provide the full table of~\autoref{new-Table-1} in the main paper in~\autoref{table_cood_ill}. Specifically, we include other neural network models beyond ResNet50 and CLIP-ResNet50. We see that except for ResNet50 on ImageNet-A, none of the other combinations of models and datasets have zero accuracies in classifying the C-OOD data. Except for the CLIP-ResNet50 which is not fine-tuned on the ImageNet data, we see that a higher ACC often leads to a lower FPR. With that being said, we argue that it may not be ideal to reject all the C-OOD examples because some of them could indeed be correctly classified.

\subsection{Tables and additional scatter plots for \autoref{ss_exp_ID}, \autoref{ss_exp_COOD}, and \autoref{ss_exp_SOOD}}
We provide in~\autoref{table_id_cood_sood} the full table used to generate \autoref{fig_id_scatter}, \autoref{fig_cood_scatter}, and \autoref{fig_sood_scatter} in the main paper. Specifically, we provide the results for each of the S-OOD datasets. It is worth noting that when we use TPR(ID$+$)=95 to select the threshold, MSP performs quite well in rejecting S-OOD data. In some datasets or when paired with some neural network models, it can even achieve the best performance (\ie, lowest FPR) compared to other detection methods. Such a superior performance degrades when we use TPR(ID)=95 to select the threshold, as will be discussed in~\autoref{suppl_ss_sood_comp} and shown in~\autoref{table_sood_comp}.

\begin{figure}
\centering
\includegraphics[width=0.87\linewidth]{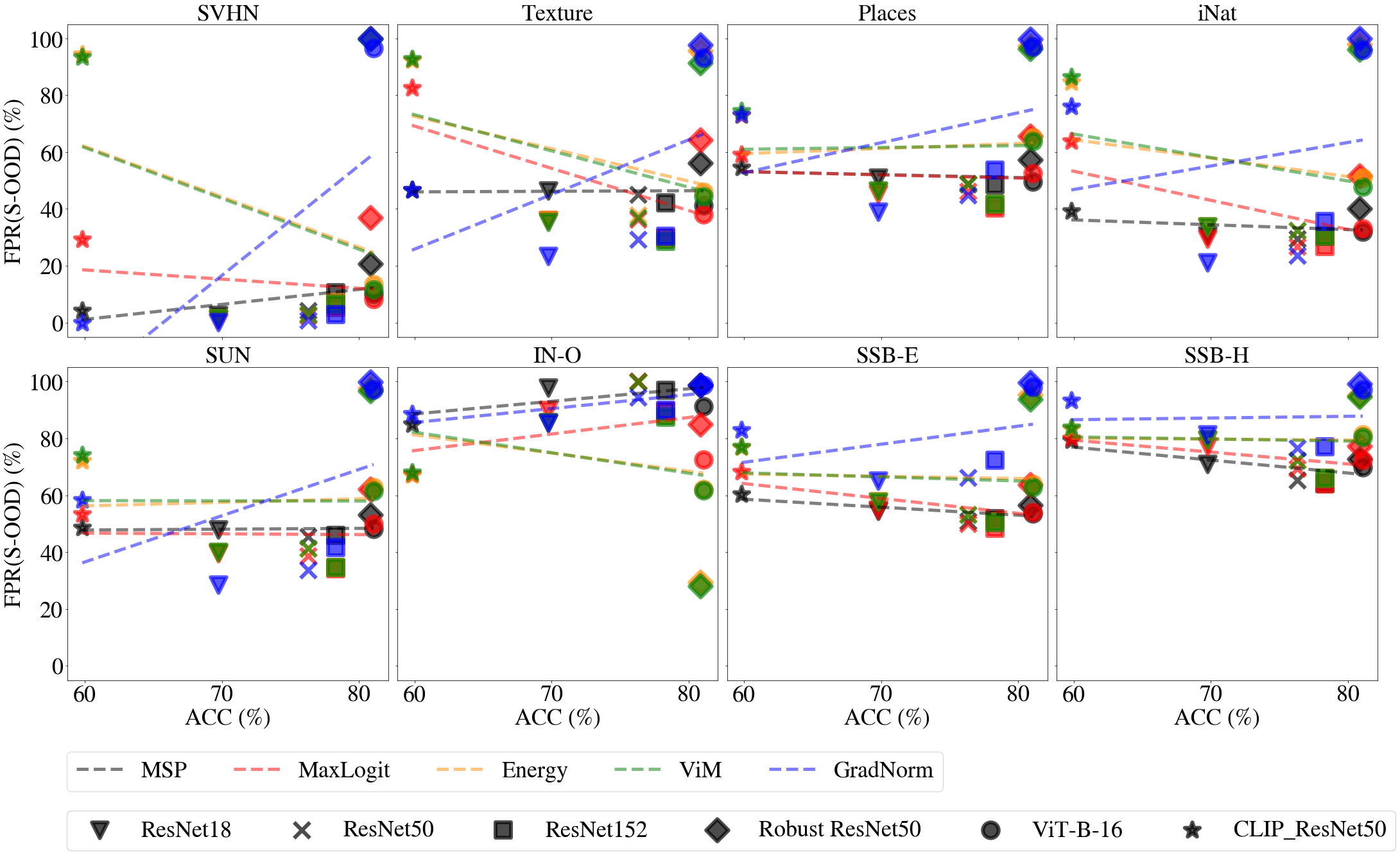}
\vskip -10pt
\caption{\small \textbf{S-OOD in \Ours for each individual dataset.} X-axis: ACC of classifying the ID examples; Y-axis: {{\color{Plum}FPR(S-OOD)}{\color{Black}@TPR(ID$+$)=95}} for wrongly accepting S-OOD examples. Please be referred to \autoref{table_id_cood_sood} for detailed values.} 

\label{fig_sood_scatter_all}
\vskip -5pt
\end{figure}

We further provide the scatter plots similar to \autoref{fig_sood_scatter} in the main text, but now for each of the eight S-OOD datasets separately in~\autoref{fig_sood_scatter_all}. The trends generally follow what we described in~\autoref{ss_exp_SOOD}. For instance, robust ResNet50, though achieving a higher ID accuracy, performs quite poorly in rejecting S-OOD data. GradNorm degrades (consistently across datasets) when the ID accuracy increases.

\begin{figure}
\centering
\includegraphics[width=0.87\linewidth]{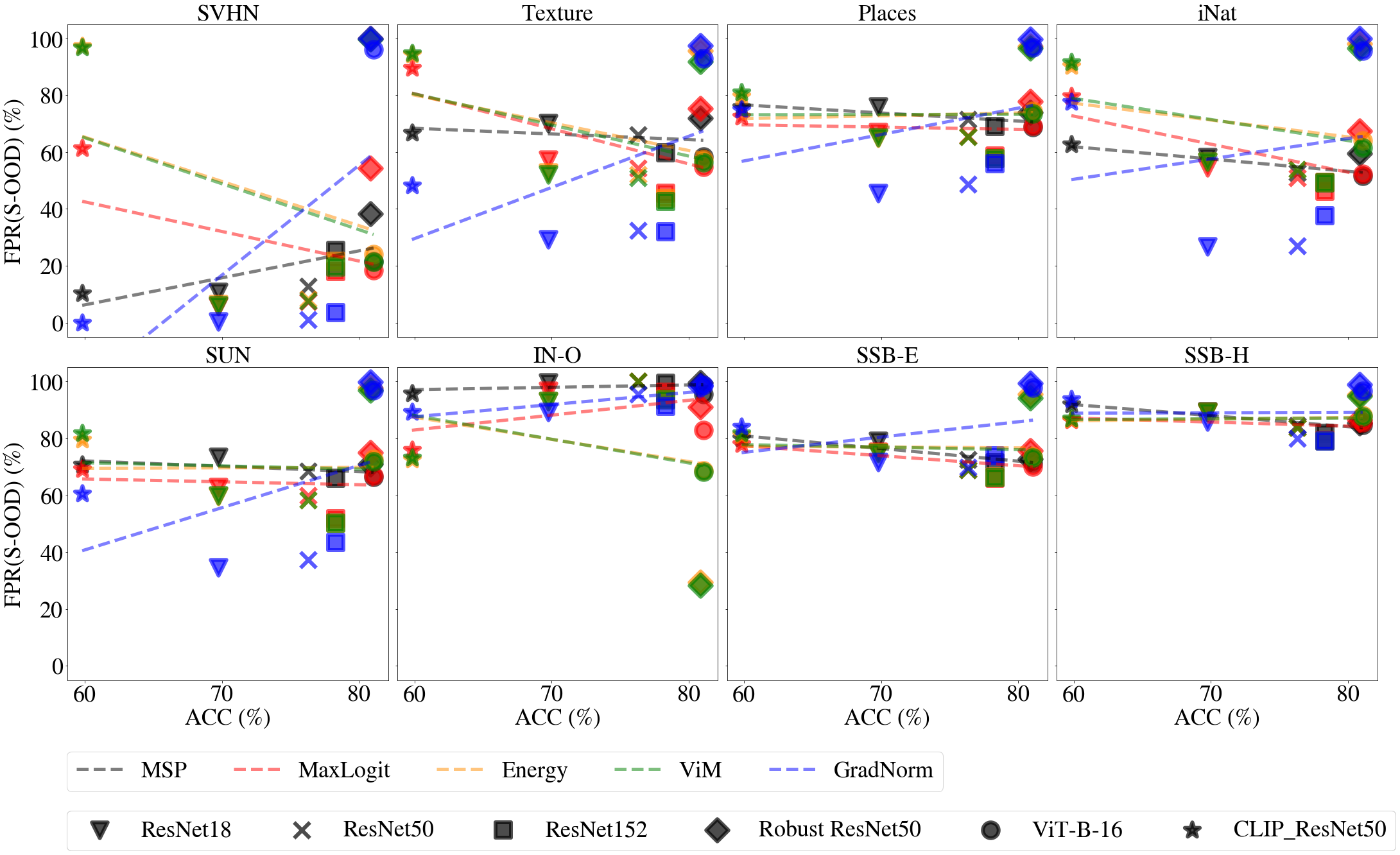}
\vskip -10pt
\caption{\small \textbf{S-OOD in conventional OOD detection framework for each individual dataset.} X-axis: ACC of classifying the ID examples; Y-axis: {{\color{Plum}FPR(S-OOD)}{\color{Black}@TPR(ID)=95}} for wrongly accepting S-OOD examples. Please be referred to \autoref{table_sood_comp} for detailed values.} 

\label{fig_conventional_sood_scatter_all}
\vskip -5pt
\end{figure}

\input{table_sood_comp}
\subsection{Additional comparisons for \autoref{ss_exp_SOOD}}
\label{suppl_ss_sood_comp}
We provide in~\autoref{table_sood_comp} the detailed results comparing \textbf{{\color{Plum}FPR(S-OOD)}{\color{Black}@TPR(ID$+$)=95}} and \textbf{{\color{Plum}FPR(S-OOD)}{\color{Black}@TPR(ID)=95}}.
The former is the metric used in \Ours for S-OOD data; the latter is the metric used in the conventional OOD detection framework. 
The main difference lies in whether we consider accepting wrongly classified ID data (please note that ID$=$ID$+\hspace{2pt}\cup\hspace{2pt}$ID$-$); please see~\autoref{ss_exp_SOOD} for some further discussions.
For the latter (\ie, \textbf{{\color{Plum}FPR(S-OOD)}{\color{Black}@TPR(ID)=95}}), we also provide the   scatter plots similar to \autoref{fig_sood_scatter} in~\autoref{fig_conventional_sood_scatter_all}.

Overall, we have three key observations. First, the trends in~\autoref{fig_conventional_sood_scatter_all} generally follow those in~\autoref{fig_sood_scatter_all}. Second, as shown in~\autoref{table_sood_comp}, using \textbf{{\color{Plum}FPR(S-OOD)}{\color{Black}@TPR(ID$+$)=95}} leads to better performance (lower FPR) in rejecting S-OOD data. Third, the baseline MSP improves the most after switching the metric from the conventional one (\ie, \textbf{{\color{Plum}FPR(S-OOD)}{\color{Black}@TPR(ID)=95}}) to the new one (\ie, \textbf{{\color{Plum}FPR(S-OOD)}{\color{Black}@TPR(ID$+$)=95}}). In some datasets or when paired with some neural network models, MSP can even achieve the best performance (\ie, lowest \textbf{{\color{Plum}FPR(S-OOD)}{\color{Black}@TPR(ID$+$)=95}}) compared to other detection methods. In other words, the poor performance of MSP in the conventional metric mainly results from the need to accommodate the misclassified ID (\ie, ID$-$) data: these examples normally have much lower MSP scores; to accept them requires a lower threshold $\tau$, hence increasing the number of wrongly accepted S-OOD examples (see~\autoref{suppl_ss_hist_all} for some further illustrations). In our metric, a detection method does not need to accept ID$-$ examples but rejects them, allowing the use of a higher threshold $\tau$ and benefiting MSP the most.

\begin{figure}
\centering
\includegraphics[width=0.75\linewidth]{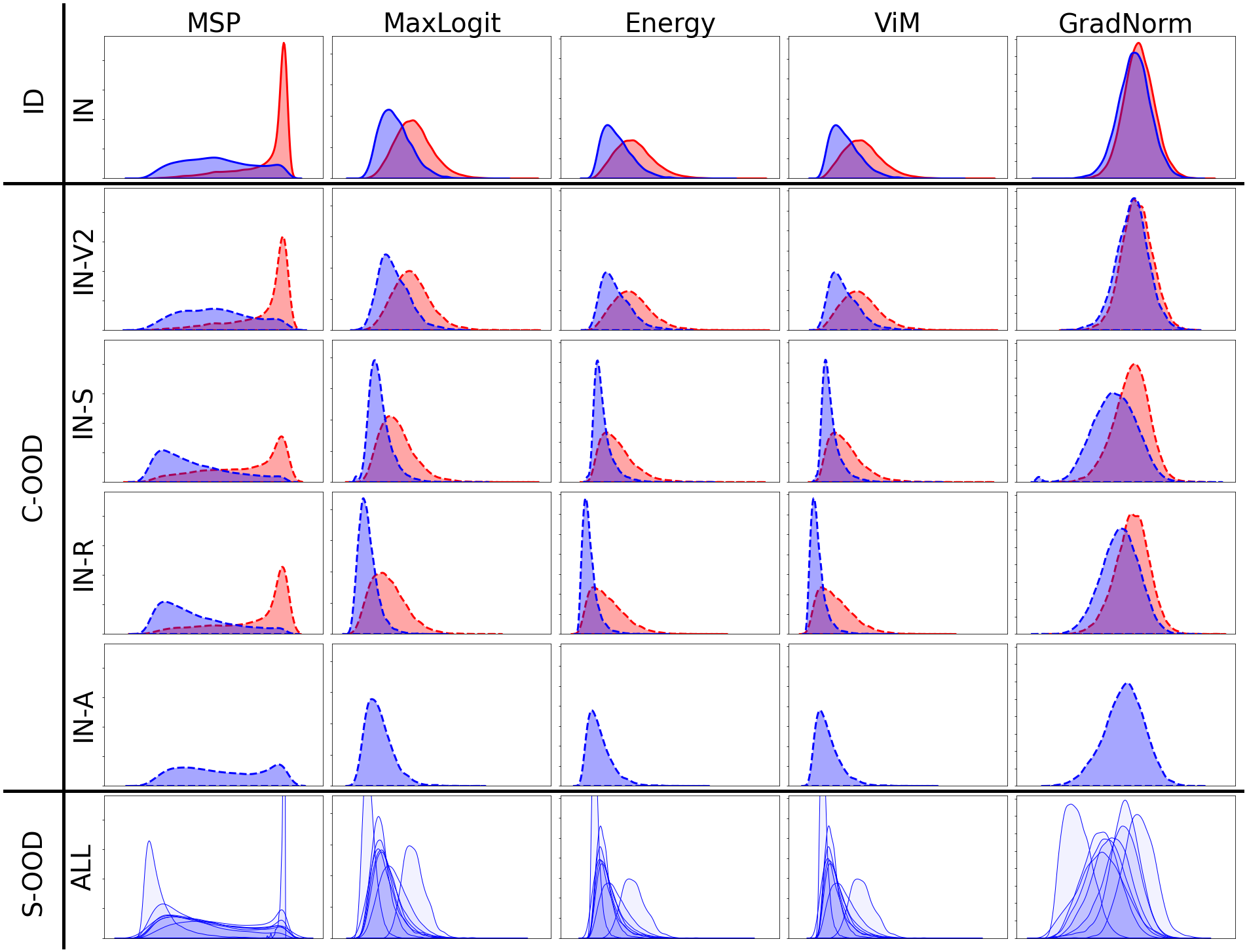}
\vskip -10pt
\caption{\small \textbf{Histogram of ID$+$, ID$-$, C-OOD$+$, C-OOD$-$ and S-OOD across all datasets at different $g(x,f)$.} We use ResNet50 as the model $f$. {\color{red}Red} and {\color{blue}Blue} region denote \textit{correctly classified} and \textit{misclassified} data respectively. The spike distribution on MSP and S-OOD pair comes from ImageNet-O \cite{imagenet_a_hendrycks2021nae} which is collected specifically to fool a ResNet50 model using MSP.} 
\label{fig_id_cood_sood_hist_resnet50}
\vskip -5pt
\end{figure}

\begin{figure}
\centering
\includegraphics[width=0.75\linewidth]{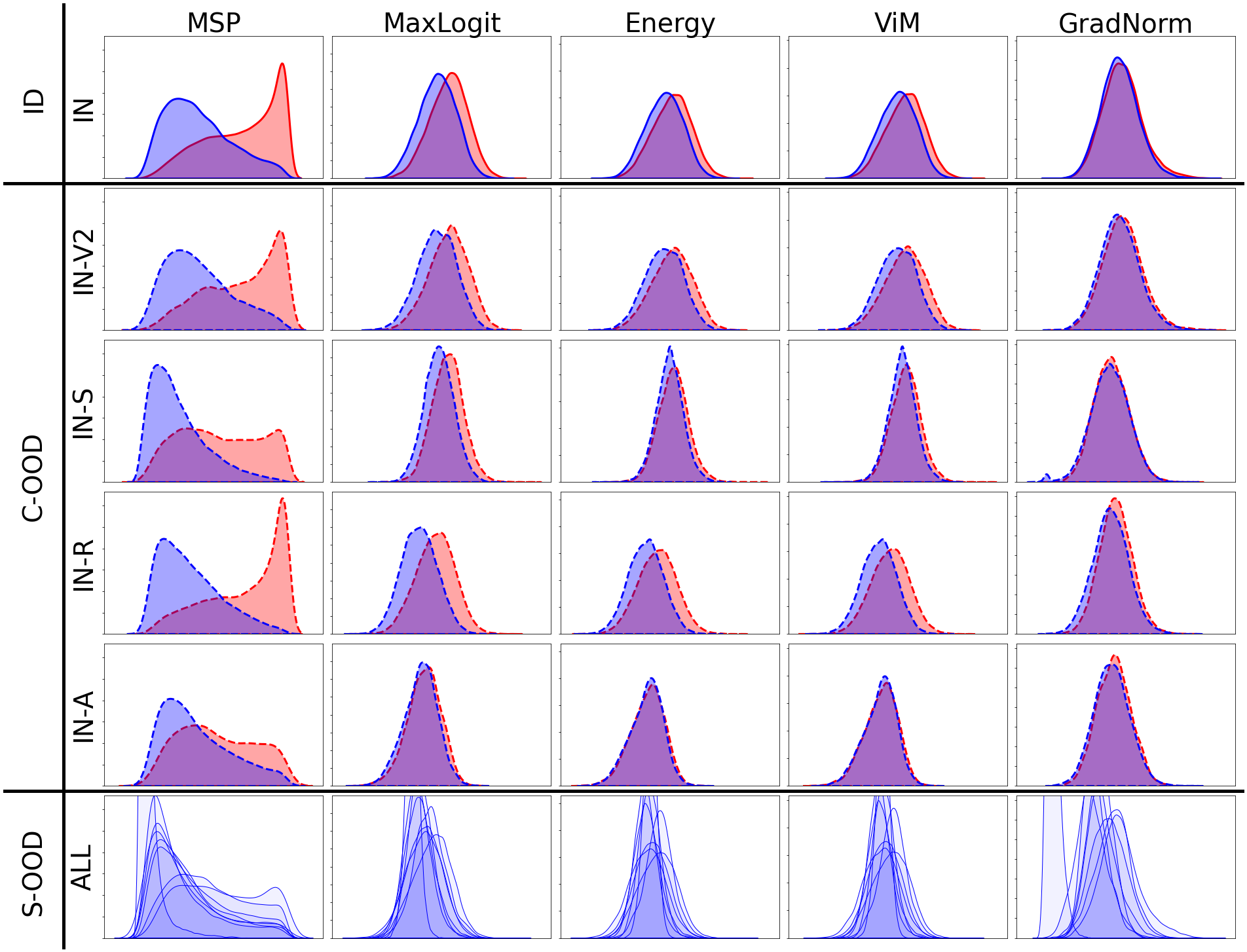}
\vskip -10pt
\caption{\small \textbf{Histogram of ID$+$, ID$-$, C-OOD$+$, C-OOD$-$ and S-OOD across all datasets at different $g(x,f)$.} We use CLIP-ResNet50 as the model $f$. {\color{red}Red} and {\color{blue}Blue} region denote \textit{correctly classified} and \textit{misclassified} data respectively.} 
\label{fig_id_cood_sood_hist_clip_resnet50}
\vskip -5pt
\end{figure}

\subsection{Additional histogram plots for \autoref{ss_exp_ID}, \autoref{ss_exp_COOD}, and \autoref{ss_exp_SOOD}}
\label{suppl_ss_hist_all}

We provide additional histogram plots in~\autoref{fig_id_cood_sood_hist_resnet50} and \autoref{fig_id_cood_sood_hist_clip_resnet50} using ResNet50 and CLIP-ResNet50, respectively. For each neural network model, we draw the $g(x, f)$ histogram for each detection method on each data type (ID, C-OOD, and S-OOD) and dataset --- we normalize separately for ID$+$, ID$-$, C-OOD$+$, C-OOD$-$, and S-OOD examples to better showcase their distribution differences. We use {\color{red}red} color to denote the acceptance cases (\ie, ID$+$ and C-OOD$+$); {\color{blue}blue} color to denote the rejection cases (\ie, ID$-$, C-OOD$-$, and S-OOD). As shown, different combinations of datasets, detection methods, and neural network models have quite different histograms. It is worth noting that for ID data (first row in each figure), MSP has the best distinction between ID$+$ and ID$-$ data: the ID$-$ examples have much lower scores (a long tail to the left). We observe a similar trend on C-OOD: MSP can best distinguish C-OOD+ and C-OOD-, with C-OOD- having lower scores.

\input{table_cood_more}
\subsection{Additional metrics for \autoref{ss_exp_COOD}}
\label{suppl_ss_cood_metrics}

We consider additional metrics (\ie, FPR(C-OOD$-$) and TPR(C-OOD$+$)) for C-OOD data in~\autoref{table_cood_more}, using the threshold selected at TPR(ID$+$)=95. Generally speaking, MSP achieves the best on both ends; \ie, low FPR and high TPR. We also observe that other methods besides MSP have impressive results on FPR(C-OOD$-$) (\ie rejecting misclassified C-OOD) but come at the cost of also rejecting most correctly classified C-OOD (\ie the TPR(C-OOD$+$) is lower). These results further illustrate why 1) using TPR(ID)=95, MSP performs poorly on FPR(S-OOD) as it needs a pretty low threshold; 2) using TPR(ID$+$)=95, MSP can choose a much larger threshold to obtain a much better FPR(S-OOD).

\subsection{Additional OOD methods for \autoref{ss_exp_ID}, \autoref{ss_exp_COOD}, and \autoref{ss_exp_SOOD}}

\begin{figure}
\centering
\includegraphics[width=1\linewidth]{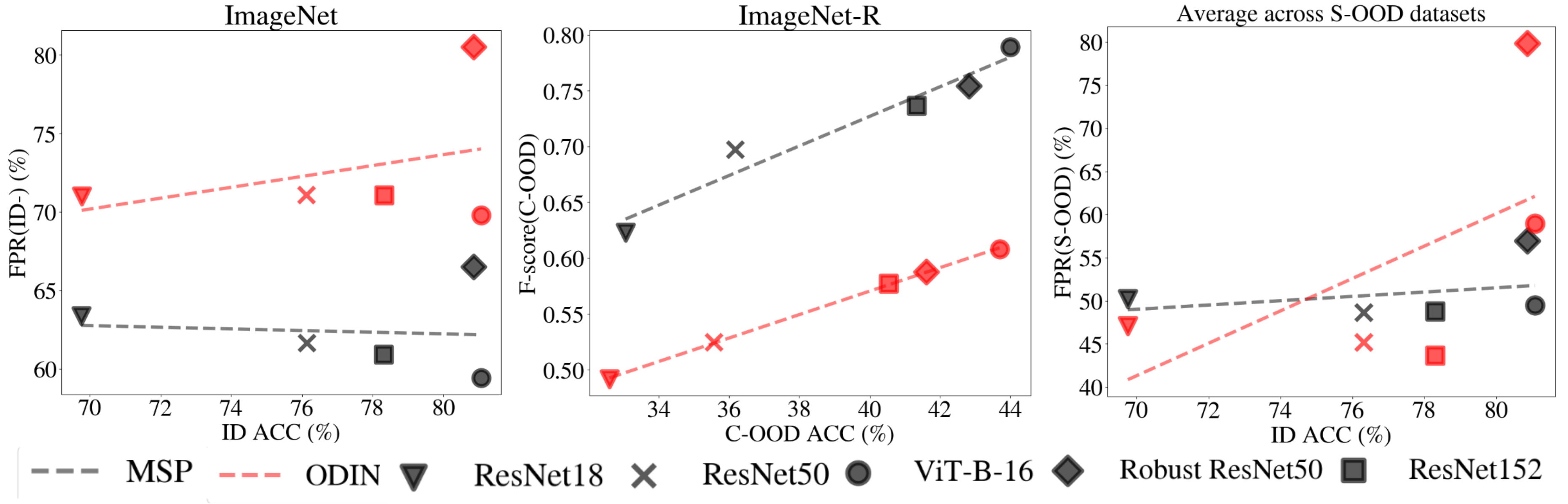}
\vskip -10pt
\caption{\small \textbf{ODIN performance in \Ours}. We use ImageNet-R for C-OOD and averaged over S-OOD datasets for S-OOD. Denotations follow \autoref{fig_id_scatter}, \autoref{fig_cood_scatter}, and \autoref{fig_sood_scatter}}
\vskip -10pt
\label{fig_odin_exp}
\end{figure}

We include ODIN in our experiment and set the hyperparameter the same as in~\cite{liang2017enhancing_odin}. We summarize our results in~\autoref{fig_odin_exp}, in which we conduct the same experiments as in~\autoref{fig_id_scatter},~\autoref{fig_cood_scatter} (ImageNet-R), and~\autoref{fig_sood_scatter}. In general, we see similar trends as MSP: on ID$-$ (left) and S-OOD (right), the higher the ID$+$ accuracy is, the lower the FPR is, except for robust ResNet. For C-OOD, the higher the C-OOD$+$ accuracy is, the higher the F-score (C-OOD).

%% file: table_cood_ill.tex
\begin{table}[]
\centering
\caption{The accuracy (ACC) to classify C-OOD data using different models, and the false positive rate (FPR) to reject C-OOD data using maximum softmax probabilities (MSP).}
\begin{tabular}{c|cc|cc|cc|cc}
\hline
\multirow{2}{*}{MODEL} & \multicolumn{2}{c|}{IN-V2}                                         & \multicolumn{2}{c|}{IN-S}                                          & \multicolumn{2}{c|}{IN-R}                                          & \multicolumn{2}{c}{IN-A}                                           \\ \cline{2-9} 
                       & ACC↑ & \begin{tabular}[c]{@{}c@{}}FPR↓\\      (C-OOD)\end{tabular} & ACC↑ & \begin{tabular}[c]{@{}c@{}}FPR↓\\      (C-OOD)\end{tabular} & ACC↑ & \begin{tabular}[c]{@{}c@{}}FPR↓\\      (C-OOD)\end{tabular} & ACC↑ & \begin{tabular}[c]{@{}c@{}}FPR↓\\      (C-OOD)\end{tabular} \\ \hline
ResNet18               & 66.5 & 94.4                                                        & 20.2 & 68.9                                                        & 33.1 & 84.5                                                        & 1.1  & 87.4                                                        \\
ResNet50               & 72.4 & 93.9                                                        & 24.1 & 65.7                                                        & 36.2 & 71.6                                                        & 0.0  & 81.5                                                        \\
ResNet152              & 75.1 & 93.7                                                        & 28.5 & 66.5                                                        & 41.3 & 70.4                                                        & 6.0  & 75.3                                                        \\
Robust   ResNet50      & 77.7 & 93.4                                                        & 29.9 & 68.8                                                        & 42.8 & 63.7                                                        & 14.5 & 74.9                                                        \\
ViT-B-16               & 77.4 & 94.2                                                        & 29.4 & 60.1                                                        & 44.0 & 54.2                                                        & 20.8 & 60.3                                                        \\
CLIP-ResNet50          & 59.5 & 95.4                                                        & 35.5 & 78.4                                                        & 60.6 & 92.3                                                        & 22.8 & 89.5  \\                                      \hline               
\end{tabular}
\label{table_cood_ill}
\end{table}

%% file: table_id_cood_sood.tex
\begin{table}[]
\scriptsize
\centering
\tabcolsep 6.5pt
\caption{\small \textbf{Performance across six models, five detection methods, and thirteen datasets under our \Ours framework.} We use FPR(ID$-$), F$_1$-score(C-OOD), and FPR(S-OOD) in percentage to evaluate each data type. The threshold is selected at TPR(ID$+$)=95. For each model, we also report its classification accuracies (ACC) on the ID and each of the C-OOD datasets. {\color{red}Red} text indicates the best-performing detection method for each combination of the neural network models and datasets.}
\begin{tabular}{c|l|c|cccc|cccccccc}
\hline
                                                                                           & \multicolumn{1}{c|}{}                         & ID                                   & \multicolumn{4}{c|}{C-OOD}                                                                                                                                & \multicolumn{8}{c}{S-OOD}                                                                                                                                                                                                                                                                                            \\ \cline{3-15} 
                                                                                           & \multicolumn{1}{c|}{}                         & FPR(ID-)↓                            & \multicolumn{4}{c|}{F-score(C-OOD)↑}                                                                                                                      & \multicolumn{8}{c}{FPR(S-OOD)↓}                                                                                                                                                                                                                                                                                      \\ \cline{3-15} 
\multirow{-3}{*}{MODEL}                                                                    & \multicolumn{1}{c|}{\multirow{-3}{*}{METHOD}} & IN                                   & IN-V2                                & IN-S                                 & IN-R                                 & IN-A                                 & SVHN                                & Texture                              & Places                               & iNat                                 & SUN                                  & IN-O                                 & SSB-E                                & SSB-H                                \\ \hline
                                                                                           & \textit{\textbf{ACC(\%)}}                     & \textit{\textbf{69.8}}               & \textit{\textbf{66.5}}               & \textit{\textbf{20.2}}               & \textit{\textbf{33.1}}               & \textit{\textbf{1.2}}                & \multicolumn{8}{c}{\textit{\textbf{$\times$}}}                                                                                                                                                                                                                                                                              \\ \cline{2-15} 
                                                                                           & MSP                                           & {\color[HTML]{FF0000} \textbf{63.4}} & {\color[HTML]{FF0000} \textbf{83.5}} & 51.5                                 & 62.3                                 & 1.7                                  & 2.5                                 & 46.3                                 & 51.0                                 & 30.7                                 & 48.0                                 & 97.8                                 & {\color[HTML]{FF0000} \textbf{54.5}} & {\color[HTML]{FF0000} \textbf{71.0}} \\
                                                                                           & MaxLogit                                      & 71.4                                 & 81.7                                 & {\color[HTML]{FF0000} \textbf{52.5}} & {\color[HTML]{FF0000} \textbf{64.6}} & 1.7                                  & 1.1                                 & 36.1                                 & 45.3                                 & 29.5                                 & 39.5                                 & 89.9                                 & 55.0                                 & 77.4                                 \\
                                                                                           & Energy                                        & 75.1                                 & 81.1                                 & 50.6                                 & 60.6                                 & 1.4                                  & 1.5                                 & 36.1                                 & 46.2                                 & 33.7                                 & 39.8                                 & 85.8                                 & 57.9                                 & 79.6                                 \\
                                                                                           & ViM                                           & 75.1                                 & 81.1                                 & 50.8                                 & 60.6                                 & 1.4                                  & 0.6                                 & 27.0                                 & 85.9                                 & 83.7                                 & 87.2                                 & {\color[HTML]{FF0000} \textbf{64.0}} & 68.9                                 & 79.6                                 \\
\multirow{-6}{*}{\textbf{ResNet18}}                                                        & GradNorm                                      & 87.5                                 & 79.0                                 & 42.9                                 & 54.0                                 & {\color[HTML]{FF0000} \textbf{2.4}}  & {\color[HTML]{FF0000} \textbf{0.1}} & {\color[HTML]{FF0000} \textbf{23.4}} & {\color[HTML]{FF0000} \textbf{38.9}} & {\color[HTML]{FF0000} \textbf{21.2}} & {\color[HTML]{FF0000} \textbf{28.6}} & 85.5                                 & 65.1                                 & 81.4                                 \\ \hline
                                                                                           & \textit{\textbf{ACC(\%)}}                     & \textit{\textbf{76.1}}               & \textit{\textbf{72.4}}               & \textit{\textbf{24.1}}               & \textit{\textbf{36.2}}               & \textit{\textbf{0.0}}                & \multicolumn{8}{c}{\textit{\textbf{$\times$}}}                                                                                                                                                                                                                                                                              \\ \cline{2-15} 
                                                                                           & MSP                                           & {\color[HTML]{FF0000} \textbf{61.7}} & {\color[HTML]{FF0000} \textbf{86.6}} & 55.7                                 & {\color[HTML]{FF0000} \textbf{69.8}} & {\color[HTML]{FF0000} \textbf{0.0}}  & 4.2                                 & 45.1                                 & 48.8                                 & 29.6                                 & 45.4                                 & 100.0                                & 50.8                                 & {\color[HTML]{FF0000} \textbf{65.2}} \\
                                                                                           & MaxLogit                                      & 70.7                                 & 85.3                                 & {\color[HTML]{FF0000} \textbf{58.4}} & 63.8                                 & {\color[HTML]{FF0000} \textbf{0.0}}  & 2.3                                 & 36.3                                 & 46.2                                 & 26.8                                 & 38.8                                 & 100.0                                & {\color[HTML]{FF0000} \textbf{50.0}} & 69.4                                 \\
                                                                                           & Energy                                        & 74.9                                 & 84.9                                 & 57.0                                 & 60.4                                 & {\color[HTML]{FF0000} \textbf{0.0}}  & 2.9                                 & 37.8                                 & 48.9                                 & 32.5                                 & 41.2                                 & 99.9                                 & 53.4                                 & 72.3                                 \\
                                                                                           & ViM                                           & 78.7                                 & 83.8                                 & 57.8                                 & 61.4                                 & {\color[HTML]{FF0000} \textbf{0.0}}  & {\color[HTML]{FF0000} \textbf{0.2}} & {\color[HTML]{FF0000} \textbf{9.1}}  & 72.4                                 & 55.5                                 & 69.7                                 & {\color[HTML]{FF0000} \textbf{79.2}} & 63.3                                 & 72.4                                 \\
\multirow{-6}{*}{\textbf{ResNet50}}                                                        & GradNorm                                      & 89.6                                 & 82.5                                 & 48.5                                 & 57.5                                 & {\color[HTML]{FF0000} \textbf{0.0}}  & 0.9                                 & 29.3                                 & {\color[HTML]{FF0000} \textbf{44.8}} & {\color[HTML]{FF0000} \textbf{23.7}} & {\color[HTML]{FF0000} \textbf{33.8}} & 94.4                                 & 66.3                                 & 76.5                                 \\ \hline
                                                                                           & \textit{\textbf{ACC(\%)}}                     & \textit{\textbf{78.3}}               & \textit{\textbf{75.1}}               & \textit{\textbf{28.5}}               & \textit{\textbf{41.3}}               & \textit{\textbf{6.0}}                & \multicolumn{8}{c}{\textit{\textbf{$\times$}}}                                                                                                                                                                                                                                                                              \\ \cline{2-15} 
                                                                                           & MSP                                           & {\color[HTML]{FF0000} \textbf{60.9}} & {\color[HTML]{FF0000} \textbf{87.9}} & 60.1                                 & {\color[HTML]{FF0000} \textbf{73.7}} & 12.9                                 & 10.5                                & 42.1                                 & 48.4                                 & 30.3                                 & 45.9                                 & 97.0                                 & 51.7                                 & 64.3                                 \\
                                                                                           & MaxLogit                                      & 70.6                                 & 86.4                                 & {\color[HTML]{FF0000} \textbf{61.7}} & 65.2                                 & {\color[HTML]{FF0000} \textbf{13.0}} & 5.4                                 & 29.6                                 & {\color[HTML]{FF0000} \textbf{40.2}} & {\color[HTML]{FF0000} \textbf{26.9}} & {\color[HTML]{FF0000} \textbf{34.1}} & 89.6                                 & {\color[HTML]{FF0000} \textbf{48.4}} & {\color[HTML]{FF0000} \textbf{64.0}} \\
                                                                                           & Energy                                        & 73.5                                 & 85.9                                 & 60.9                                 & 62.6                                 & 13.0                                 & 7.2                                 & 29.5                                 & 41.7                                 & 30.7                                 & 34.7                                 & 88.1                                 & 50.6                                 & 66.0                                 \\
                                                                                           & ViM                                           & 77.8                                 & 85.9                                 & 61.5                                 & 65.3                                 & 9.8                                  & {\color[HTML]{FF0000} \textbf{0.1}} & {\color[HTML]{FF0000} \textbf{7.7}}  & 64.7                                 & 44.0                                 & 63.4                                 & {\color[HTML]{FF0000} \textbf{61.2}} & 60.2                                 & 66.1                                 \\
\multirow{-6}{*}{\textbf{ResNet152}}                                                       & GradNorm                                      & 91.7                                 & 84.0                                 & 51.2                                 & 60.8                                 & 12.2                                 & 2.9                                 & 30.3                                 & 53.7                                 & 35.5                                 & 41.7                                 & 89.9                                 & 72.3                                 & 77.1                                 \\ \hline
                                                                                           & \textit{\textbf{ACC(\%)}}                     & \textit{\textbf{80.9}}               & \textit{\textbf{77.7}}               & \textit{\textbf{29.9}}               & \textit{\textbf{42.8}}               & \textit{\textbf{14.6}}               & \multicolumn{8}{c}{\textit{\textbf{$\times$}}}                                                                                                                                                                                                                                                                              \\ \cline{2-15} 
                                                                                           & MSP                                           & {\color[HTML]{FF0000} \textbf{66.5}} & {\color[HTML]{FF0000} \textbf{88.4}} & {\color[HTML]{FF0000} \textbf{59.4}} & {\color[HTML]{FF0000} \textbf{75.4}} & {\color[HTML]{FF0000} \textbf{28.2}} & 20.8                                & 56.2                                 & {\color[HTML]{FF0000} \textbf{57.3}} & 40.2                                 & {\color[HTML]{FF0000} \textbf{53.1}} & 98.8                                 & {\color[HTML]{FF0000} \textbf{56.6}} & {\color[HTML]{FF0000} \textbf{72.8}} \\
                                                                                           & MaxLogit                                      & 72.2                                 & 87.8                                 & 56.2                                 & 68.6                                 & 23.5                                 & 37.0                                & 64.3                                 & 65.5                                 & 51.5                                 & 62.2                                 & 85.0                                 & 63.8                                 & 77.3                                 \\
                                                                                           & Energy                                        & 94.4                                 & 85.6                                 & 45.2                                 & 20.7                                 & 7.2                                  & 100.0                               & 95.6                                 & 97.2                                 & 98.0                                 & 97.6                                 & {\color[HTML]{FF0000} \textbf{29.4}} & 95.5                                 & 95.2                                 \\
                                                                                           & ViM                                           & 85.1                                 & 86.5                                 & 59.3                                 & 63.3                                 & 20.6                                 & {\color[HTML]{FF0000} \textbf{0.0}} & {\color[HTML]{FF0000} \textbf{18.7}} & 72.3                                 & {\color[HTML]{FF0000} \textbf{25.5}} & 67.3                                 & 57.6                                 & 67.4                                 & 94.6                                 \\
\multirow{-6}{*}{\textbf{\begin{tabular}[c]{@{}c@{}}Robust \\      ResNet50\end{tabular}}} & GradNorm                                      & 99.2                                 & 85.0                                 & 45.8                                 & 59.8                                 & 25.4                                 & 100.0                               & 97.8                                 & 99.8                                 & 100.0                                & 99.8                                 & 98.8                                 & 99.6                                 & 99.1                                 \\ \hline
                                                                                           & \textit{\textbf{ACC(\%)}}                     & \textit{\textbf{81.1}}               & \textit{\textbf{77.4}}               & \textit{\textbf{29.4}}               & \textit{\textbf{44.0}}               & \textit{\textbf{20.8}}               & \multicolumn{8}{c}{\textit{\textbf{$\times$}}}                                                                                                                                                                                                                                                                              \\ \cline{2-15} 
                                                                                           & MSP                                           & {\color[HTML]{FF0000} \textbf{59.4}} & {\color[HTML]{FF0000} \textbf{88.9}} & 63.7                                 & {\color[HTML]{FF0000} \textbf{78.9}} & {\color[HTML]{FF0000} \textbf{38.8}} & 10.1                                & 41.3                                 & {\color[HTML]{FF0000} \textbf{49.7}} & 32.0                                 & {\color[HTML]{FF0000} \textbf{48.1}} & 91.3                                 & {\color[HTML]{FF0000} \textbf{53.5}} & {\color[HTML]{FF0000} \textbf{69.8}} \\
                                                                                           & MaxLogit                                      & 63.9                                 & 88.3                                 & {\color[HTML]{FF0000} \textbf{64.0}} & 69.8                                 & 29.8                                 & 8.2                                 & 38.1                                 & 52.5                                 & 33.0                                 & 49.9                                 & 72.6                                 & 54.1                                 & 72.3                                 \\
                                                                                           & Energy                                        & 75.8                                 & 87.0                                 & 62.3                                 & 61.1                                 & 25.5                                 & 13.5                                & 46.0                                 & 65.2                                 & 50.9                                 & 62.8                                 & 62.2                                 & 64.0                                 & 81.3                                 \\
                                                                                           & ViM                                           & 76.3                                 & 87.0                                 & 60.1                                 & 59.6                                 & 14.3                                 & {\color[HTML]{FF0000} \textbf{0.8}} & {\color[HTML]{FF0000} \textbf{31.5}} & 50.4                                 & {\color[HTML]{FF0000} \textbf{10.7}} & 49.1                                 & {\color[HTML]{FF0000} \textbf{61.8}} & 58.7                                 & 80.4                                 \\
\multirow{-6}{*}{\textbf{ViT-B-16}}                                                        & GradNorm                                      & 98.1                                 & 84.6                                 & 46.9                                 & 61.3                                 & 34.7                                 & 96.4                                & 93.1                                 & 96.9                                 & 95.9                                 & 97.1                                 & 98.8                                 & 98.0                                 & 97.1                                 \\ \hline
                                                                                           & \textit{\textbf{ACC(\%)}}                     & \textit{\textbf{59.8}}               & \textit{\textbf{59.5}}               & \textit{\textbf{35.5}}               & \textit{\textbf{60.6}}               & \textit{\textbf{22.8}}               & \multicolumn{8}{c}{\textit{\textbf{$\times$}}}                                                                                                                                                                                                                                                                              \\ \cline{2-15} 
                                                                                           & MSP                                           & {\color[HTML]{FF0000} \textbf{72.1}} & {\color[HTML]{FF0000} \textbf{77.5}} & {\color[HTML]{FF0000} \textbf{63.1}} & {\color[HTML]{FF0000} \textbf{80.3}} & {\color[HTML]{FF0000} \textbf{41.3}} & 4.2                                 & {\color[HTML]{FF0000} \textbf{46.6}} & {\color[HTML]{FF0000} \textbf{54.6}} & {\color[HTML]{FF0000} \textbf{39.3}} & {\color[HTML]{FF0000} \textbf{48.7}} & 85.0                                 & {\color[HTML]{FF0000} \textbf{60.3}} & {\color[HTML]{FF0000} \textbf{79.2}} \\
                                                                                           & MaxLogit                                      & 86.6                                 & 74.5                                 & 53.1                                 & 76.0                                 & 36.4                                 & 29.3                                & 82.6                                 & 59.1                                 & 63.8                                 & 53.5                                 & 67.4                                 & 68.3                                 & 80.0                                 \\
                                                                                           & Energy                                        & 90.7                                 & 73.9                                 & 52.3                                 & 73.6                                 & 34.9                                 & 94.2                                & 92.3                                 & 72.8                                 & 84.6                                 & 72.2                                 & {\color[HTML]{FF0000} \textbf{67.1}} & 76.7                                 & 83.4                                 \\
                                                                                           & ViM                                           & 90.7                                 & 73.8                                 & 52.3                                 & 73.3                                 & 34.8                                 & 93.4                                & 92.7                                 & 74.3                                 & 86.3                                 & 74.1                                 & 68.1                                 & 77.0                                 & 83.7                                 \\
\multirow{-6}{*}{\textbf{\begin{tabular}[c]{@{}c@{}}CLIP\\      ResNet50\end{tabular}}}    & GradNorm                                      & 93.9                                 & 73.4                                 & 51.5                                 & 74.9                                 & 37.5                                 & {\color[HTML]{FF0000} \textbf{0.0}} & 46.7                                 & 73.2                                 & 75.9                                 & 58.4                                 & 88.5                                 & 83.0                                 & 93.3                     \\ \hline           
\end{tabular}
\label{table_id_cood_sood}
\end{table}

%% file: table_sood_comp.tex
\begin{table}[]
\scriptsize
\centering
\tabcolsep 6pt
\caption{\textbf{S-OOD performance comparison between conventional framework and \Ours}. Conventional framework uses TPR(ID)=95 (left column) while \Ours uses TPR(ID$+$)=95 (right column). {\color{red}Red} indicates the best value for a particular dataset and framework.}
\begin{tabular}{c|l|cccccccccccccclc}
\hline
                                                                                           & \multicolumn{1}{c|}{}                         & \multicolumn{16}{c}{FPR(S-OOD)↓}                                                                                                                                                                                                                                                                                                                                                                                                                                                                                                                                                                                                                                                                                                                                                                 \\ \cline{3-18} 
                                                                                           & \multicolumn{1}{c|}{}                         & \multicolumn{16}{c}{TPR(ID) vs TPR(ID+)}                                                                                                                                                                                                                                                                                                                                                                                                                                                                                                                                                                                                                                                                                                                                                        \\ \cline{3-18} 
\multirow{-3}{*}{MODEL}                                                                    & \multicolumn{1}{c|}{\multirow{-3}{*}{METHOD}} & \multicolumn{2}{c|}{SVHN}                                                                      & \multicolumn{2}{c|}{Texture}                                                                     & \multicolumn{2}{c|}{Places}                                                                      & \multicolumn{2}{c|}{iNat}                                                                        & \multicolumn{2}{c|}{SUN}                                                                         & \multicolumn{2}{c|}{IN-O}                                                                        & \multicolumn{2}{c|}{SSB-E}                                                                       & \multicolumn{2}{c}{SSB-H}                                                    \\ \hline
                                                                                           & MSP                                           & 10.7                                & \multicolumn{1}{c|}{2.5}                                 & 70.2                                 & \multicolumn{1}{c|}{46.3}                                 & 75.9                                 & \multicolumn{1}{c|}{51.0}                                 & 58.2                                 & \multicolumn{1}{c|}{30.7}                                 & 73.5                                 & \multicolumn{1}{c|}{48.0}                                 & 99.7                                 & \multicolumn{1}{c|}{97.8}                                 & 79.0                                 & \multicolumn{1}{c|}{{\color[HTML]{FF0000} \textbf{54.5}}} & 89.38                                 & {\color[HTML]{FF0000} \textbf{71.0}} \\
                                                                                           & MaxLogit                                      & 5.8                                 & \multicolumn{1}{c|}{1.1}                                 & 57.2                                 & \multicolumn{1}{c|}{36.1}                                 & 66.9                                 & \multicolumn{1}{c|}{45.3}                                 & 54.5                                 & \multicolumn{1}{c|}{29.5}                                 & 62.9                                 & \multicolumn{1}{c|}{39.5}                                 & 96.8                                 & \multicolumn{1}{c|}{89.9}                                 & 75.4                                 & \multicolumn{1}{c|}{55.0}                                 & 89.24                                 & 77.4                                 \\
                                                                                           & Energy                                        & 6.5                                 & \multicolumn{1}{c|}{1.5}                                 & 52.8                                 & \multicolumn{1}{c|}{36.1}                                 & 65.0                                 & \multicolumn{1}{c|}{46.2}                                 & 56.5                                 & \multicolumn{1}{c|}{33.7}                                 & 59.8                                 & \multicolumn{1}{c|}{39.8}                                 & 93.2                                 & \multicolumn{1}{c|}{85.8}                                 & 75.0                                 & \multicolumn{1}{c|}{57.9}                                 & 89.26                                 & 79.6                                 \\
                                                                                           & ViM                                           & 1.6                                 & \multicolumn{1}{c|}{0.6}                                 & 38.8                                 & \multicolumn{1}{c|}{27.0}                                 & 92.5                                 & \multicolumn{1}{c|}{85.9}                                 & 91.2                                 & \multicolumn{1}{c|}{83.7}                                 & 93.1                                 & \multicolumn{1}{c|}{87.2}                                 & {\color[HTML]{FF0000} \textbf{72.7}} & \multicolumn{1}{c|}{{\color[HTML]{FF0000} \textbf{64.0}}} & 80.1                                 & \multicolumn{1}{c|}{68.9}                                 & 89.26                                 & 79.6                                 \\
\multirow{-5}{*}{\textbf{ResNet18}}                                                        & GradNorm                                      & {\color[HTML]{FF0000} \textbf{0.3}} & \multicolumn{1}{c|}{{\color[HTML]{FF0000} \textbf{0.1}}} & {\color[HTML]{FF0000} \textbf{29.3}} & \multicolumn{1}{c|}{{\color[HTML]{FF0000} \textbf{23.4}}} & {\color[HTML]{FF0000} \textbf{45.5}} & \multicolumn{1}{c|}{{\color[HTML]{FF0000} \textbf{38.9}}} & {\color[HTML]{FF0000} \textbf{26.8}} & \multicolumn{1}{c|}{{\color[HTML]{FF0000} \textbf{21.2}}} & {\color[HTML]{FF0000} \textbf{34.7}} & \multicolumn{1}{c|}{{\color[HTML]{FF0000} \textbf{28.6}}} & 89.3                                 & \multicolumn{1}{c|}{85.5}                                 & {\color[HTML]{FF0000} \textbf{71.7}} & \multicolumn{1}{c|}{65.1}                                 & {\color[HTML]{FF0000} \textbf{85.92}} & 81.4                                 \\ \hline
                                                                                           & MSP                                           & 12.9                                & \multicolumn{1}{c|}{4.2}                                 & 66.0                                 & \multicolumn{1}{c|}{45.1}                                 & 71.6                                 & \multicolumn{1}{c|}{48.8}                                 & 52.8                                 & \multicolumn{1}{c|}{29.6}                                 & 68.6                                 & \multicolumn{1}{c|}{45.4}                                 & 100.0                                & \multicolumn{1}{c|}{100.0}                                & 72.6                                 & \multicolumn{1}{c|}{50.8}                                 & 84.53                                 & {\color[HTML]{FF0000} \textbf{65.2}} \\
                                                                                           & MaxLogit                                      & 7.5                                 & \multicolumn{1}{c|}{2.3}                                 & 54.4                                 & \multicolumn{1}{c|}{36.3}                                 & 65.7                                 & \multicolumn{1}{c|}{46.2}                                 & 50.9                                 & \multicolumn{1}{c|}{26.8}                                 & 59.9                                 & \multicolumn{1}{c|}{38.8}                                 & 100.0                                & \multicolumn{1}{c|}{100.0}                                & {\color[HTML]{FF0000} \textbf{69.0}} & \multicolumn{1}{c|}{{\color[HTML]{FF0000} \textbf{50.0}}} & 83.76                                 & 69.4                                 \\
                                                                                           & Energy                                        & 8.2                                 & \multicolumn{1}{c|}{2.9}                                 & 52.1                                 & \multicolumn{1}{c|}{37.8}                                 & 65.4                                 & \multicolumn{1}{c|}{48.9}                                 & 54.0                                 & \multicolumn{1}{c|}{32.5}                                 & 58.3                                 & \multicolumn{1}{c|}{41.2}                                 & 100.0                                & \multicolumn{1}{c|}{99.9}                                 & 69.2                                 & \multicolumn{1}{c|}{53.4}                                 & 83.87                                 & 72.3                                 \\
                                                                                           & ViM                                           & {\color[HTML]{FF0000} \textbf{0.8}} & \multicolumn{1}{c|}{{\color[HTML]{FF0000} \textbf{0.2}}} & {\color[HTML]{FF0000} \textbf{15.7}} & \multicolumn{1}{c|}{{\color[HTML]{FF0000} \textbf{9.1}}}  & 83.5                                 & \multicolumn{1}{c|}{72.4}                                 & 71.8                                 & \multicolumn{1}{c|}{55.5}                                 & 82.1                                 & \multicolumn{1}{c|}{69.7}                                 & {\color[HTML]{FF0000} \textbf{84.9}} & \multicolumn{1}{c|}{{\color[HTML]{FF0000} \textbf{79.2}}} & 76.2                                 & \multicolumn{1}{c|}{63.3}                                 & 83.92                                 & 72.4                                 \\
\multirow{-5}{*}{\textbf{ResNet50}}                                                        & GradNorm                                      & 1.1                                 & \multicolumn{1}{c|}{0.9}                                 & 32.4                                 & \multicolumn{1}{c|}{29.3}                                 & {\color[HTML]{FF0000} \textbf{48.7}} & \multicolumn{1}{c|}{{\color[HTML]{FF0000} \textbf{44.8}}} & {\color[HTML]{FF0000} \textbf{27.0}} & \multicolumn{1}{c|}{{\color[HTML]{FF0000} \textbf{23.7}}} & {\color[HTML]{FF0000} \textbf{37.3}} & \multicolumn{1}{c|}{{\color[HTML]{FF0000} \textbf{33.8}}} & 95.6                                 & \multicolumn{1}{c|}{94.4}                                 & 69.8                                 & \multicolumn{1}{c|}{66.3}                                 & {\color[HTML]{FF0000} \textbf{80.06}} & 76.5                                 \\ \hline
                                                                                           & MSP                                           & 25.5                                & \multicolumn{1}{c|}{10.5}                                & 59.8                                 & \multicolumn{1}{c|}{42.1}                                 & 68.9                                 & \multicolumn{1}{c|}{48.4}                                 & 49.3                                 & \multicolumn{1}{c|}{30.3}                                 & 66.0                                 & \multicolumn{1}{c|}{45.9}                                 & 99.3                                 & \multicolumn{1}{c|}{97.0}                                 & 71.3                                 & \multicolumn{1}{c|}{51.7}                                 & 81.88                                 & 64.3                                 \\
                                                                                           & MaxLogit                                      & 18.0                                & \multicolumn{1}{c|}{5.4}                                 & 45.6                                 & \multicolumn{1}{c|}{29.6}                                 & 58.7                                 & \multicolumn{1}{c|}{{\color[HTML]{FF0000} \textbf{40.2}}} & 46.3                                 & \multicolumn{1}{c|}{{\color[HTML]{FF0000} \textbf{26.9}}} & 51.9                                 & \multicolumn{1}{c|}{{\color[HTML]{FF0000} \textbf{34.1}}} & 96.2                                 & \multicolumn{1}{c|}{89.6}                                 & {\color[HTML]{FF0000} \textbf{66.1}} & \multicolumn{1}{c|}{{\color[HTML]{FF0000} \textbf{48.4}}} & 79.4                                  & {\color[HTML]{FF0000} \textbf{64.0}} \\
                                                                                           & Energy                                        & 21.5                                & \multicolumn{1}{c|}{7.2}                                 & 43.8                                 & \multicolumn{1}{c|}{29.5}                                 & 57.7                                 & \multicolumn{1}{c|}{41.7}                                 & 49.4                                 & \multicolumn{1}{c|}{30.7}                                 & 50.3                                 & \multicolumn{1}{c|}{34.7}                                 & 93.7                                 & \multicolumn{1}{c|}{88.1}                                 & 66.4                                 & \multicolumn{1}{c|}{50.6}                                 & {\color[HTML]{FF0000} \textbf{79.01}} & 66.0                                 \\
                                                                                           & ViM                                           & {\color[HTML]{FF0000} \textbf{0.3}} & \multicolumn{1}{c|}{{\color[HTML]{FF0000} \textbf{0.1}}} & {\color[HTML]{FF0000} \textbf{13.0}} & \multicolumn{1}{c|}{{\color[HTML]{FF0000} \textbf{7.7}}}  & 78.6                                 & \multicolumn{1}{c|}{64.7}                                 & 63.1                                 & \multicolumn{1}{c|}{44.0}                                 & 77.8                                 & \multicolumn{1}{c|}{63.4}                                 & {\color[HTML]{FF0000} \textbf{71.5}} & \multicolumn{1}{c|}{{\color[HTML]{FF0000} \textbf{61.2}}} & 73.5                                 & \multicolumn{1}{c|}{60.2}                                 & 79.2                                  & 66.1                                 \\
\multirow{-5}{*}{\textbf{ResNet152}}                                                       & GradNorm                                      & 3.5                                 & \multicolumn{1}{c|}{2.9}                                 & 31.9                                 & \multicolumn{1}{c|}{30.3}                                 & {\color[HTML]{FF0000} \textbf{55.9}} & \multicolumn{1}{c|}{53.7}                                 & {\color[HTML]{FF0000} \textbf{37.5}} & \multicolumn{1}{c|}{35.5}                                 & {\color[HTML]{FF0000} \textbf{43.5}} & \multicolumn{1}{c|}{41.7}                                 & 91.3                                 & \multicolumn{1}{c|}{89.9}                                 & 74.0                                 & \multicolumn{1}{c|}{72.3}                                 & 79.12                                 & 77.1                                 \\ \hline
                                                                                           & MSP                                           & 38.4                                & \multicolumn{1}{c|}{20.8}                                & 71.9                                 & \multicolumn{1}{c|}{56.2}                                 & {\color[HTML]{FF0000} \textbf{74.1}} & \multicolumn{1}{c|}{{\color[HTML]{FF0000} \textbf{57.3}}} & 59.5                                 & \multicolumn{1}{c|}{40.2}                                 & {\color[HTML]{FF0000} \textbf{70.9}} & \multicolumn{1}{c|}{{\color[HTML]{FF0000} \textbf{53.1}}} & 99.6                                 & \multicolumn{1}{c|}{98.8}                                 & {\color[HTML]{FF0000} \textbf{72.8}} & \multicolumn{1}{c|}{{\color[HTML]{FF0000} \textbf{56.6}}} & {\color[HTML]{FF0000} \textbf{85.44}} & {\color[HTML]{FF0000} \textbf{72.8}} \\
                                                                                           & MaxLogit                                      & 54.3                                & \multicolumn{1}{c|}{37.0}                                & 75.4                                 & \multicolumn{1}{c|}{64.3}                                 & 77.8                                 & \multicolumn{1}{c|}{65.5}                                 & 67.4                                 & \multicolumn{1}{c|}{51.5}                                 & 75.0                                 & \multicolumn{1}{c|}{62.2}                                 & 91.0                                 & \multicolumn{1}{c|}{85.0}                                 & 75.7                                 & \multicolumn{1}{c|}{63.8}                                 & 86.48                                 & 77.3                                 \\
                                                                                           & Energy                                        & 100.0                               & \multicolumn{1}{c|}{100.0}                               & 95.7                                 & \multicolumn{1}{c|}{95.6}                                 & 97.3                                 & \multicolumn{1}{c|}{97.2}                                 & 98.1                                 & \multicolumn{1}{c|}{98.0}                                 & 97.7                                 & \multicolumn{1}{c|}{97.6}                                 & {\color[HTML]{FF0000} \textbf{29.5}} & \multicolumn{1}{c|}{{\color[HTML]{FF0000} \textbf{29.4}}} & 95.6                                 & \multicolumn{1}{c|}{95.5}                                 & 95.28                                 & 95.2                                 \\
                                                                                           & ViM                                           & {\color[HTML]{FF0000} \textbf{0.1}} & \multicolumn{1}{c|}{{\color[HTML]{FF0000} \textbf{0.0}}} & {\color[HTML]{FF0000} \textbf{21.9}} & \multicolumn{1}{c|}{{\color[HTML]{FF0000} \textbf{18.7}}} & 77.6                                 & \multicolumn{1}{c|}{72.3}                                 & {\color[HTML]{FF0000} \textbf{31.4}} & \multicolumn{1}{c|}{{\color[HTML]{FF0000} \textbf{25.5}}} & 73.3                                 & \multicolumn{1}{c|}{67.3}                                 & 64.7                                 & \multicolumn{1}{c|}{57.6}                                 & 73.9                                 & \multicolumn{1}{c|}{67.4}                                 & 94.97                                 & 94.6                                 \\
\multirow{-5}{*}{\textbf{\begin{tabular}[c]{@{}c@{}}Robust \\      ResNet50\end{tabular}}} & GradNorm                                      & 100.0                               & \multicolumn{1}{c|}{100.0}                               & 97.4                                 & \multicolumn{1}{c|}{97.8}                                 & 99.6                                 & \multicolumn{1}{c|}{99.8}                                 & 100.0                                & \multicolumn{1}{c|}{100.0}                                & 99.8                                 & \multicolumn{1}{c|}{99.8}                                 & 98.6                                 & \multicolumn{1}{c|}{98.8}                                 & 99.5                                 & \multicolumn{1}{c|}{99.6}                                 & 98.89                                 & 99.1                                 \\ \hline
                                                                                           & MSP                                           & 21.5                                & \multicolumn{1}{c|}{10.1}                                & 58.3                                 & \multicolumn{1}{c|}{41.3}                                 & 68.7                                 & \multicolumn{1}{c|}{{\color[HTML]{FF0000} \textbf{49.7}}} & 51.5                                 & \multicolumn{1}{c|}{32.0}                                 & 66.6                                 & \multicolumn{1}{c|}{{\color[HTML]{FF0000} \textbf{48.1}}} & 95.7                                 & \multicolumn{1}{c|}{91.3}                                 & 71.3                                 & \multicolumn{1}{c|}{{\color[HTML]{FF0000} \textbf{53.5}}} & {\color[HTML]{FF0000} \textbf{84.73}} & {\color[HTML]{FF0000} \textbf{69.8}} \\
                                                                                           & MaxLogit                                      & 18.5                                & \multicolumn{1}{c|}{8.2}                                 & 54.8                                 & \multicolumn{1}{c|}{38.1}                                 & 69.1                                 & \multicolumn{1}{c|}{52.5}                                 & 52.3                                 & \multicolumn{1}{c|}{33.0}                                 & 66.9                                 & \multicolumn{1}{c|}{49.9}                                 & 82.9                                 & \multicolumn{1}{c|}{72.6}                                 & {\color[HTML]{FF0000} \textbf{70.2}} & \multicolumn{1}{c|}{54.1}                                 & 85.64                                 & 72.3                                 \\
                                                                                           & Energy                                        & 24.1                                & \multicolumn{1}{c|}{13.5}                                & 57.4                                 & \multicolumn{1}{c|}{46.0}                                 & 74.3                                 & \multicolumn{1}{c|}{65.2}                                 & 64.1                                 & \multicolumn{1}{c|}{50.9}                                 & 72.8                                 & \multicolumn{1}{c|}{62.8}                                 & {\color[HTML]{FF0000} \textbf{68.7}} & \multicolumn{1}{c|}{62.2}                                 & 73.8                                 & \multicolumn{1}{c|}{64.0}                                 & 88.16                                 & 81.3                                 \\
                                                                                           & ViM                                           & {\color[HTML]{FF0000} \textbf{2.3}} & \multicolumn{1}{c|}{{\color[HTML]{FF0000} \textbf{0.8}}} & {\color[HTML]{FF0000} \textbf{43.9}} & \multicolumn{1}{c|}{{\color[HTML]{FF0000} \textbf{31.5}}} & {\color[HTML]{FF0000} \textbf{61.1}} & \multicolumn{1}{c|}{50.4}                                 & {\color[HTML]{FF0000} \textbf{17.8}} & \multicolumn{1}{c|}{{\color[HTML]{FF0000} \textbf{10.7}}} & {\color[HTML]{FF0000} \textbf{59.5}} & \multicolumn{1}{c|}{49.1}                                 & 72.9                                 & \multicolumn{1}{c|}{{\color[HTML]{FF0000} \textbf{61.8}}} & 71.6                                 & \multicolumn{1}{c|}{58.7}                                 & 87.73                                 & 80.4                                 \\
\multirow{-5}{*}{\textbf{ViT-B-16}}                                                        & GradNorm                                      & 96.1                                & \multicolumn{1}{c|}{96.4}                                & 92.9                                 & \multicolumn{1}{c|}{93.1}                                 & 96.7                                 & \multicolumn{1}{c|}{96.9}                                 & 95.6                                 & \multicolumn{1}{c|}{95.9}                                 & 96.9                                 & \multicolumn{1}{c|}{97.1}                                 & 98.7                                 & \multicolumn{1}{c|}{98.8}                                 & 97.9                                 & \multicolumn{1}{c|}{98.0}                                 & 96.77                                 & 97.1                                 \\ \hline
                                                                                           & MSP                                           & 10.3                                & \multicolumn{1}{c|}{4.2}                                 & 66.8                                 & \multicolumn{1}{c|}{{\color[HTML]{FF0000} \textbf{46.6}}} & 75.9                                 & \multicolumn{1}{c|}{{\color[HTML]{FF0000} \textbf{54.6}}} & {\color[HTML]{FF0000} \textbf{62.7}} & \multicolumn{1}{c|}{{\color[HTML]{FF0000} \textbf{39.3}}} & 70.8                                 & \multicolumn{1}{c|}{{\color[HTML]{FF0000} \textbf{48.7}}} & 95.8                                 & \multicolumn{1}{c|}{85.0}                                 & 80.1                                 & \multicolumn{1}{c|}{{\color[HTML]{FF0000} \textbf{60.3}}} & 91.87                                 & {\color[HTML]{FF0000} \textbf{79.2}} \\
                                                                                           & MaxLogit                                      & 61.2                                & \multicolumn{1}{c|}{29.3}                                & 89.5                                 & \multicolumn{1}{c|}{82.6}                                 & {\color[HTML]{FF0000} \textbf{72.3}} & \multicolumn{1}{c|}{59.1}                                 & 79.5                                 & \multicolumn{1}{c|}{63.8}                                 & 69.1                                 & \multicolumn{1}{c|}{53.5}                                 & 75.9                                 & \multicolumn{1}{c|}{67.4}                                 & {\color[HTML]{FF0000} \textbf{77.9}} & \multicolumn{1}{c|}{68.3}                                 & {\color[HTML]{FF0000} \textbf{86.25}} & 80.0                                 \\
                                                                                           & Energy                                        & 97.1                                & \multicolumn{1}{c|}{94.2}                                & 94.4                                 & \multicolumn{1}{c|}{92.3}                                 & 79.5                                 & \multicolumn{1}{c|}{72.8}                                 & 90.1                                 & \multicolumn{1}{c|}{84.6}                                 & 79.6                                 & \multicolumn{1}{c|}{72.2}                                 & {\color[HTML]{FF0000} \textbf{72.7}} & \multicolumn{1}{c|}{{\color[HTML]{FF0000} \textbf{67.1}}} & 81.2                                 & \multicolumn{1}{c|}{76.7}                                 & 86.59                                 & 83.4                                 \\
                                                                                           & ViM                                           & 96.7                                & \multicolumn{1}{c|}{93.4}                                & 94.7                                 & \multicolumn{1}{c|}{92.7}                                 & 80.9                                 & \multicolumn{1}{c|}{74.3}                                 & 91.6                                 & \multicolumn{1}{c|}{86.3}                                 & 81.8                                 & \multicolumn{1}{c|}{74.1}                                 & 73.3                                 & \multicolumn{1}{c|}{68.1}                                 & 81.7                                 & \multicolumn{1}{c|}{77.0}                                 & 87.03                                 & 83.7                                 \\
\multirow{-5}{*}{\textbf{\begin{tabular}[c]{@{}c@{}}CLIP\\      ResNet50\end{tabular}}}    & GradNorm                                      & {\color[HTML]{FF0000} \textbf{0.0}} & \multicolumn{1}{c|}{{\color[HTML]{FF0000} \textbf{0.0}}} & {\color[HTML]{FF0000} \textbf{48.3}} & \multicolumn{1}{c|}{46.7}                                 & 75.0                                 & \multicolumn{1}{c|}{73.2}                                 & 77.5                                 & \multicolumn{1}{c|}{75.9}                                 & {\color[HTML]{FF0000} \textbf{60.6}} & \multicolumn{1}{c|}{58.4}                                 & 89.2                                 & \multicolumn{1}{c|}{88.5}                                 & 84.0                                 & \multicolumn{1}{c|}{83.0}                                 & 93.75                                 & 93.3                                 \\ \hline
\end{tabular}
\label{table_sood_comp}
\end{table}

%% file: table_cood_more.tex
\begin{table}[]
\scriptsize
\centering
\caption{\textbf{Extension of \autoref{table_id_cood_sood} on C-OOD datasets} including FPR(C-OOD$-$) and TPR(C-OOD$+$) defined in \autoref{suppl_s_metric}.}
\begin{tabular}{c|l|cccc|cccc}
\hline
                                                                                           & \multicolumn{1}{c|}{}                         & \multicolumn{4}{c|}{FPR(C-OOD)↓}                                                                                                                          & \multicolumn{4}{c}{TPR(C-OOD)↑}                                                                                                                            \\ \cline{3-10} 
\multirow{-2}{*}{MODEL}                                                                    & \multicolumn{1}{c|}{\multirow{-2}{*}{METHOD}} & IN-V2                                & IN-S                                 & IN-R                                 & IN-A                                 & IN-V2                                & IN-S                                 & IN-R                                 & IN-A                                  \\ \hline
                                                                                           & \textit{\textbf{ACC(\%)}}                     & \textit{\textbf{66.5}}               & \textit{\textbf{20.2}}               & \textit{\textbf{33.1}}               & \textit{\textbf{1.2}}                & \textit{\textbf{66.5}}               & \textit{\textbf{20.2}}               & \textit{\textbf{33.1}}               & \textit{\textbf{1.2}}                 \\ \cline{2-10} 
                                                                                           & MSP                                           & {\color[HTML]{FF0000} \textbf{62.3}} & 34.2                                 & 48.2                                 & 61.9                                 & {\color[HTML]{FF0000} \textbf{94.1}} & 81.3                                 & {\color[HTML]{FF0000} \textbf{89.4}} & 46.5                                  \\
                                                                                           & MaxLogit                                      & 70.8                                 & {\color[HTML]{FF0000} \textbf{28.9}} & 9.2                                  & 26.9                                 & 93.7                                 & 76.2                                 & 56.7                                 & 20.9                                  \\
                                                                                           & Energy                                        & 74.2                                 & 31.0                                 & 7.0                                  & 23.0                                 & 93.7                                 & 75.2                                 & 49.7                                 & 15.1                                  \\
                                                                                           & ViM                                           & 74.1                                 & 30.6                                 & {\color[HTML]{FF0000} \textbf{6.9}}  & {\color[HTML]{FF0000} \textbf{22.9}} & 93.7                                 & 75.0                                 & 49.5                                 & 15.1                                  \\
\multirow{-6}{*}{\textbf{ResNet18}}                                                        & GradNorm                                      & 85.5                                 & 51.1                                 & 67.8                                 & 63.2                                 & 93.5                                 & {\color[HTML]{FF0000} \textbf{82.2}} & 87.8                                 & {\color[HTML]{FF0000} \textbf{68.6}}  \\ \hline
                                                                                           & \textit{\textbf{ACC(\%)}}                     & \textit{\textbf{72.4}}               & \textit{\textbf{24.1}}               & \textit{\textbf{36.2}}               & \textit{\textbf{0.0}}                & \textit{\textbf{72.4}}               & \textit{\textbf{24.1}}               & \textit{\textbf{36.2}}               & \textit{\textbf{0.0}}                 \\ \cline{2-10} 
                                                                                           & MSP                                           & {\color[HTML]{FF0000} \textbf{60.5}} & 34.1                                 & 34.2                                 & 60.6                                 & {\color[HTML]{FF0000} \textbf{94.1}} & 80.1                                 & 85.8                                 & {\color[HTML]{FF0000} \textbf{0.0}}   \\
                                                                                           & MaxLogit                                      & 67.9                                 & 24.6                                 & 5.0                                  & 27.5                                 & 93.7                                 & 73.2                                 & 51.0                                 & {\color[HTML]{FF0000} \textbf{0.0}}   \\
                                                                                           & Energy                                        & 71.7                                 & 25.9                                 & 4.2                                  & {\color[HTML]{FF0000} \textbf{24.7}} & 93.9                                 & 72.4                                 & 46.4                                 & {\color[HTML]{FF0000} \textbf{0.0}}   \\
                                                                                           & ViM                                           & 79.7                                 & {\color[HTML]{FF0000} \textbf{17.8}} & {\color[HTML]{FF0000} \textbf{4.2}}  & 28.5                                 & 94.1                                 & 63.5                                 & 47.5                                 & {\color[HTML]{FF0000} \textbf{0.0}}   \\
\multirow{-6}{*}{\textbf{ResNet50}}                                                        & GradNorm                                      & 88.5                                 & 51.8                                 & 64.1                                 & 70.9                                 & 93.8                                 & {\color[HTML]{FF0000} \textbf{84.3}} & {\color[HTML]{FF0000} \textbf{86.0}} & {\color[HTML]{FF0000} \textbf{0.0}}   \\ \hline
                                                                                           & \textit{\textbf{ACC(\%)}}                     & \textit{\textbf{75.1}}               & \textit{\textbf{28.5}}               & \textit{\textbf{41.3}}               & \textit{\textbf{6.0}}                & \textit{\textbf{75.1}}               & \textit{\textbf{28.5}}               & \textit{\textbf{41.3}}               & \textit{\textbf{6.0}}                 \\ \cline{2-10} 
                                                                                           & MSP                                           & {\color[HTML]{FF0000} \textbf{59.6}} & 35.7                                 & 33.6                                 & 53.9                                 & 94.0                                 & 81.4                                 & 86.2                                 & 65.0                                  \\
                                                                                           & MaxLogit                                      & 68.9                                 & 26.7                                 & 4.9                                  & 20.9                                 & 93.5                                 & 74.4                                 & 51.8                                 & 29.6                                  \\
                                                                                           & Energy                                        & 72.0                                 & 27.7                                 & {\color[HTML]{FF0000} \textbf{4.3}}  & {\color[HTML]{FF0000} \textbf{19.1}} & 93.2                                 & 74.1                                 & 48.4                                 & 27.7                                  \\
                                                                                           & ViM                                           & 77.3                                 & {\color[HTML]{FF0000} \textbf{19.3}} & 4.4                                  & 23.2                                 & {\color[HTML]{FF0000} \textbf{94.6}} & 65.9                                 & 51.4                                 & 23.7                                  \\
\multirow{-6}{*}{\textbf{ResNet152}}                                                       & GradNorm                                      & 90.8                                 & 61.8                                 & 76.4                                 & 80.6                                 & 94.2                                 & {\color[HTML]{FF0000} \textbf{87.6}} & {\color[HTML]{FF0000} \textbf{90.9}} & {\color[HTML]{FF0000} \textbf{88.3}}  \\ \hline
                                                                                           & \textit{\textbf{ACC(\%)}}                     & \textit{\textbf{77.7}}               & \textit{\textbf{29.9}}               & \textit{\textbf{42.8}}               & \textit{\textbf{14.6}}               & \textit{\textbf{77.7}}               & \textit{\textbf{29.9}}               & \textit{\textbf{42.8}}               & \textit{\textbf{14.6}}                \\ \cline{2-10} 
                                                                                           & MSP                                           & {\color[HTML]{FF0000} \textbf{64.2}} & 40.0                                 & 30.3                                 & 57.5                                 & 93.7                                 & 81.8                                 & 85.0                                 & 72.0                                  \\
                                                                                           & MaxLogit                                      & 69.5                                 & 48.9                                 & 6.3                                  & 19.8                                 & 93.9                                 & 83.9                                 & 56.6                                 & 28.8                                  \\
                                                                                           & Energy                                        & 94.0                                 & 95.4                                 & {\color[HTML]{FF0000} \textbf{0.8}}  & {\color[HTML]{FF0000} \textbf{3.2}}  & {\color[HTML]{FF0000} \textbf{95.0}} & 94.6                                 & 11.6                                 & 4.4                                   \\
                                                                                           & ViM                                           & 85.3                                 & {\color[HTML]{FF0000} \textbf{39.1}} & 6.3                                  & 22.4                                 & 94.9                                 & 80.8                                 & 50.2                                 & 26.6                                  \\
\multirow{-6}{*}{\textbf{\begin{tabular}[c]{@{}c@{}}Robust \\      ResNet50\end{tabular}}} & GradNorm                                      & 99.2                                 & 100.0                                & 100.0                                & 99.9                                 & 94.9                                 & {\color[HTML]{FF0000} \textbf{99.3}} & {\color[HTML]{FF0000} \textbf{99.7}} & {\color[HTML]{FF0000} \textbf{100.0}} \\ \hline
                                                                                           & \textit{\textbf{ACC(\%)}}                     & \textit{\textbf{77.4}}               & \textit{\textbf{29.4}}               & \textit{\textbf{44.0}}               & \textit{\textbf{20.8}}               & \textit{\textbf{77.4}}               & \textit{\textbf{29.4}}               & \textit{\textbf{44.0}}               & \textit{\textbf{20.8}}                \\ \cline{2-10} 
                                                                                           & MSP                                           & {\color[HTML]{FF0000} \textbf{60.6}} & 30.4                                 & 19.0                                 & 40.3                                 & 94.3                                 & 80.8                                 & 81.0                                 & 60.8                                  \\
                                                                                           & MaxLogit                                      & 64.4                                 & 29.2                                 & 4.3                                  & 13.7                                 & 94.0                                 & 80.2                                 & 56.6                                 & 26.7                                  \\
                                                                                           & Energy                                        & 75.2                                 & 35.3                                 & {\color[HTML]{FF0000} \textbf{2.9}}  & 9.3                                  & 94.0                                 & 83.7                                 & 45.6                                 & 19.8                                  \\
                                                                                           & ViM                                           & 76.2                                 & {\color[HTML]{FF0000} \textbf{23.6}} & 2.9                                  & {\color[HTML]{FF0000} \textbf{9.2}}  & 94.2                                 & 67.3                                 & 44.0                                 & 10.4                                  \\
\multirow{-6}{*}{\textbf{ViT-B-16}}                                                        & GradNorm                                      & 98.5                                 & 92.7                                 & 94.5                                 & 96.9                                 & {\color[HTML]{FF0000} \textbf{94.4}} & {\color[HTML]{FF0000} \textbf{98.9}} & {\color[HTML]{FF0000} \textbf{97.3}} & {\color[HTML]{FF0000} \textbf{98.2}}  \\ \hline
                                                                                           & \textit{\textbf{ACC(\%)}}                     & \textit{\textbf{59.5}}               & \textit{\textbf{35.5}}               & \textit{\textbf{60.6}}               & \textit{\textbf{22.8}}               & \textit{\textbf{59.5}}               & \textit{\textbf{35.5}}               & \textit{\textbf{60.6}}               & \textit{\textbf{22.8}}                \\ \cline{2-10} 
                                                                                           & MSP                                           & {\color[HTML]{FF0000} \textbf{73.0}} & {\color[HTML]{FF0000} \textbf{45.1}} & 61.7                                 & 67.4                                 & 94.8                                 & 83.9                                 & {\color[HTML]{FF0000} \textbf{93.9}} & 85.5                                  \\
                                                                                           & MaxLogit                                      & 86.7                                 & 94.3                                 & {\color[HTML]{FF0000} \textbf{57.3}} & {\color[HTML]{FF0000} \textbf{64.4}} & 94.4                                 & 98.2                                 & 84.1                                 & 70.8                                  \\
                                                                                           & Energy                                        & 90.3                                 & 98.8                                 & 65.4                                 & 69.9                                 & 94.6                                 & {\color[HTML]{FF0000} \textbf{99.0}} & 83.0                                 & 71.2                                  \\
                                                                                           & ViM                                           & 90.5                                 & 98.5                                 & 65.6                                 & 71.2                                 & 94.5                                 & 98.8                                 & 82.6                                 & 72.0                                  \\
\multirow{-6}{*}{\textbf{\begin{tabular}[c]{@{}c@{}}CLIP\\      ResNet50\end{tabular}}}    & GradNorm                                      & 93.6                                 & 84.2                                 & 85.9                                 & 87.7                                 & {\color[HTML]{FF0000} \textbf{95.0}} & 87.6                                 & 93.4                                 & {\color[HTML]{FF0000} \textbf{91.7}} 
\\ \hline
\end{tabular}
\label{table_cood_more}
\end{table}